%% file: arxiv_paper_and_supplementary.tex
\documentclass[letterpaper, 10 pt, journal, twoside]{IEEEtran}

\usepackage[table,xcdraw,dvipsnames]{xcolor}
\usepackage[hyphens]{url}
\usepackage[hyphenbreaks]{breakurl}
\usepackage{hyperref}
\hypersetup{
           breaklinks=true,   
           colorlinks=true,   
           pdfusetitle=true,  
           linkcolor=blue,
           citecolor=blue,
           urlcolor=blue,
           hyperfigures=true
        }
\usepackage{mathptmx} 
\usepackage{amsmath} 
\usepackage{amssymb}  
\usepackage{multirow}
\usepackage{subfig}
\usepackage[titlenumbered,ruled]{algorithm2e}

\usepackage{rotating}
\usepackage{tabularx} 
\usepackage{graphicx,wrapfig}
\usepackage{caption}
\usepackage{makecell}
\usepackage{svg}
\usepackage{tabularray}

\ifCLASSINFOpdf
\else
\fi

\let\oldtwocolumn\twocolumn
\renewcommand\twocolumn[1][]{%
    \oldtwocolumn[{#1}{
    \begin{center}
		\vspace{-0.6cm}
		\resizebox{.99\textwidth}{!}{%
		\includegraphics[height=3cm]{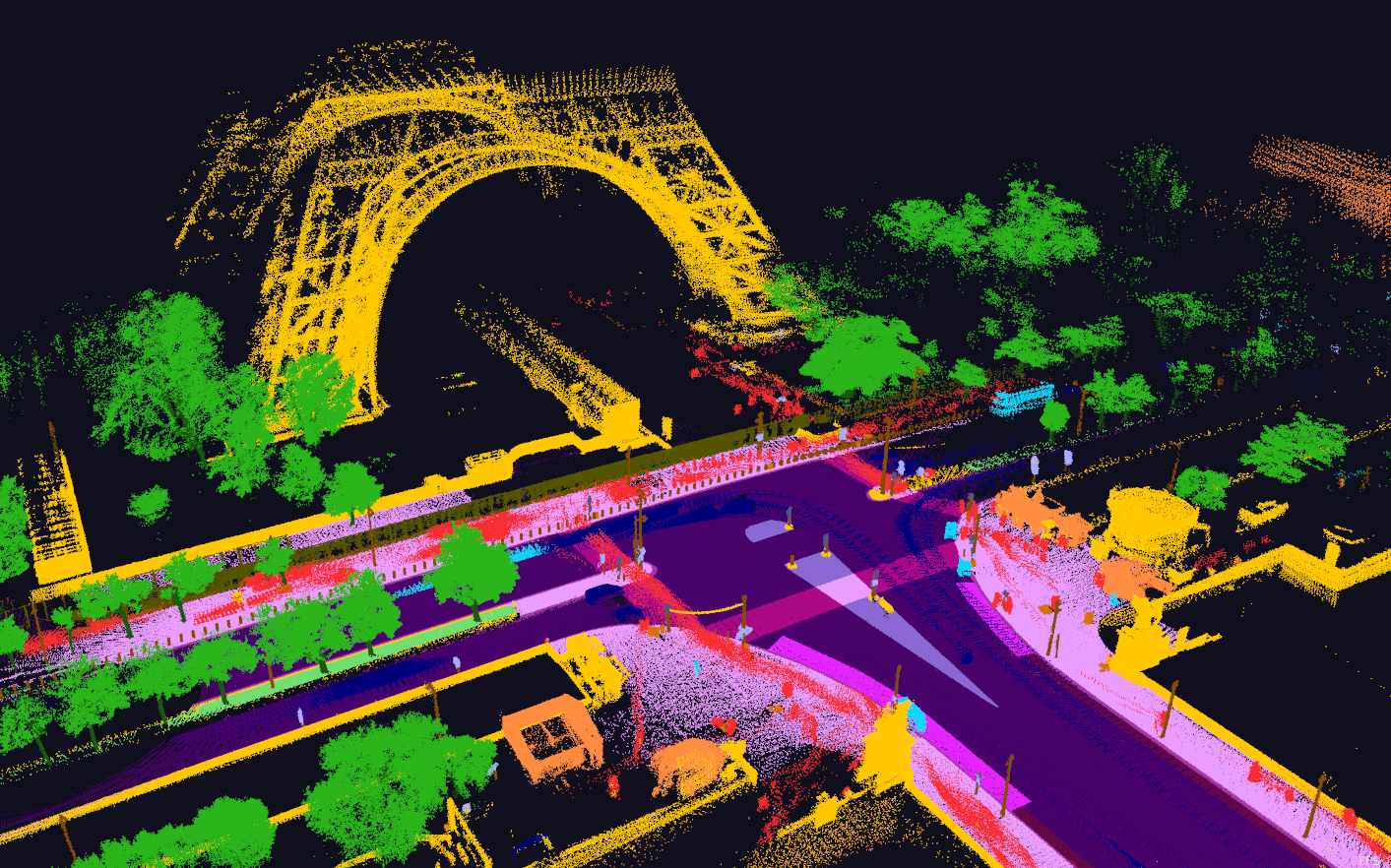}
        \includegraphics[height=3cm]{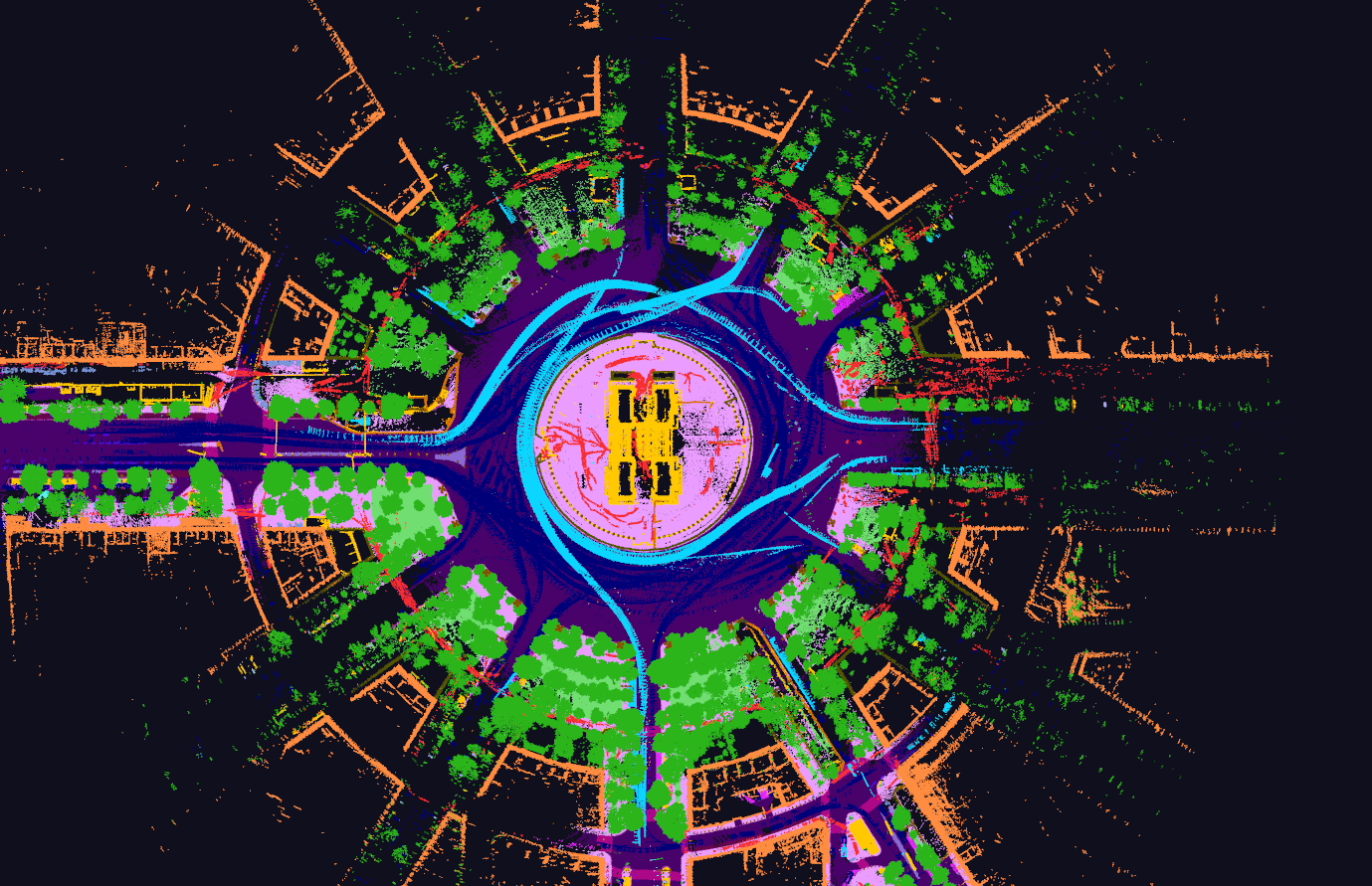}
        \includegraphics[height=3cm]{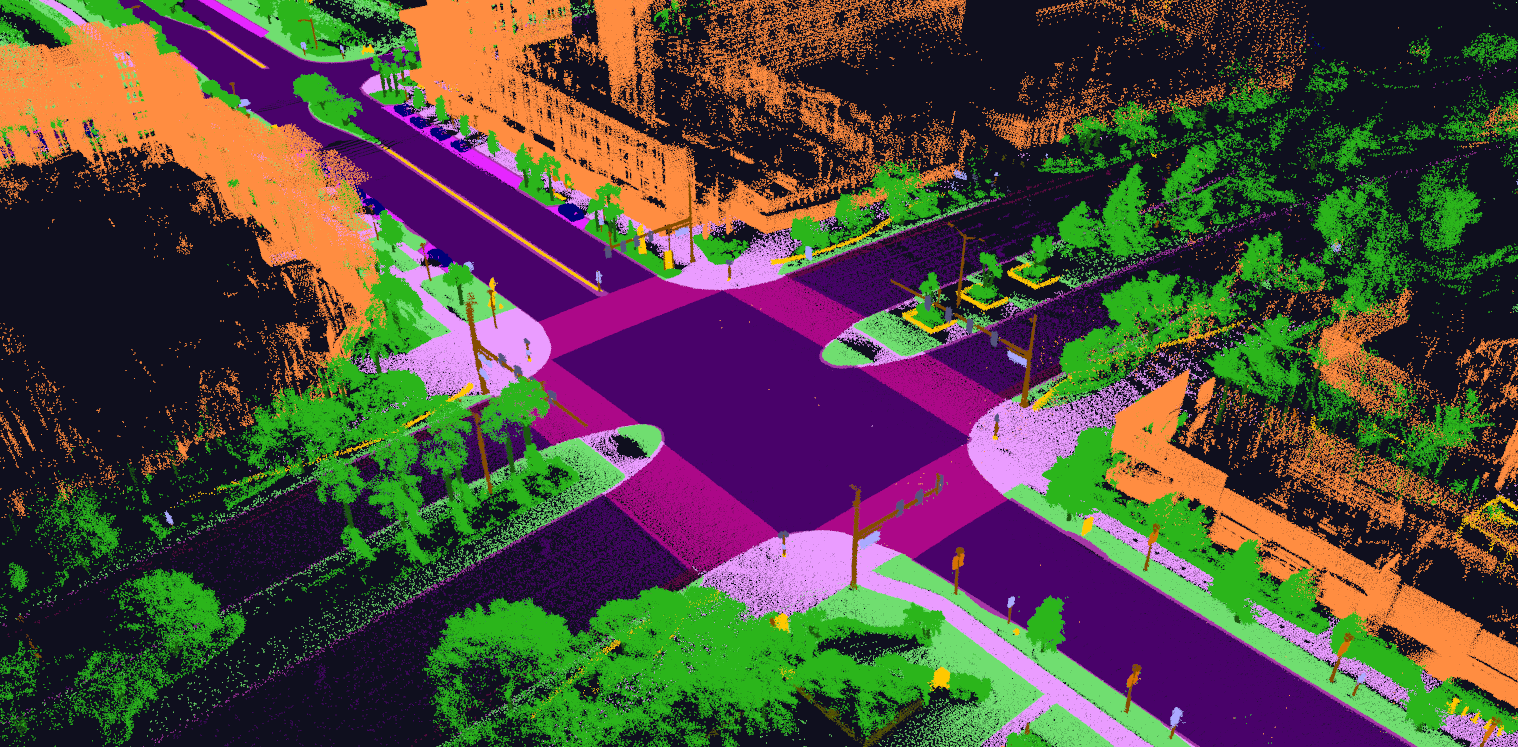}
        \vspace{-0.65cm}
        }
		\captionof{figure}
		{
			Navya3DSeg dataset has been collected across 13 different countries, with a high diversity on recorded areas.
		}
		\label{fig:n3ds_samples}
	\end{center}
    }]
}
\begin{document}
\bstctlcite{IEEEexample:BSTcontrol}
\title{\LARGE \bf Navya3DSeg - Navya 3D Semantic Segmentation \\ Dataset Design \& split generation for autonomous vehicles\\
}
%
%
%


\author{Alexandre Almin$^{1}$, Léo Lemarié$^{1}$, Anh Duong$^{1}$, B Ravi Kiran$^{1}$%
\thanks{Manuscript received: February 14, 2023; Revised May 9 2023; Accepted June 12 2023.}
\thanks{This paper was recommended for publication by
Editor Cesar Cadena upon evaluation of the Associate Editor and Reviewers’
comments.} 
\thanks{$^{1}$All authors are from Navya, Machine Learning {\tt\small firstname.lastname@navya.tech}}%
\thanks{Digital Object Identifier (DOI): see top of this page.}
}

\markboth{IEEE Robotics and Automation Letters. Accepted Version. June, 2023}
{Almin \MakeLowercase{\textit{et al.}}: Navya3DSeg} 

%



\maketitle

\begin{abstract}
Autonomous driving (AD) perception today relies heavily on deep learning based architectures requiring large scale annotated datasets with their associated costs for curation and annotation. The 3D semantic data are useful for core perception tasks such as obstacle detection and ego-vehicle localization. We propose a new dataset, Navya 3D Segmentation (Navya3DSeg), with a diverse label space corresponding to a large scale production grade operational domain, including rural, urban, industrial sites and universities from 13 countries. It contains 23 labeled sequences and 25 supplementary sequences without labels, designed to explore self-supervised and semi-supervised semantic segmentation benchmarks on point clouds. We also propose a novel method for sequential dataset split generation based on iterative multi-label stratification, and demonstrated to achieve a +1.2\% \textit{mIoU} improvement over the original split proposed by SemanticKITTI dataset. A complete benchmark for semantic segmentation task was performed, with state of the art methods. Finally, we demonstrate an Active Learning (AL) based dataset distillation framework. We introduce a novel heuristic-free sampling method
called ego-pose distance based sampling in the context of AL. A detailed presentation on the dataset is available here \url{https://www.youtube.com/watch?v=5m6ALIs-s20}.
\end{abstract}


%
\IEEEpeerreviewmaketitle



\section{INTRODUCTION}
Large scale semantic segmentation of point clouds has received lots of attention due to their utility in modern autonomous driving perception systems for creation of HDMaps, ground truth generation (3D detection, panoptic segmentation) and automated labeling systems, robustifying localization modules for Autonomous Vehicles (AV), road surface reconstruction and model fitting \cite{liu2020high}. Several models dedicated to semantic segmentation task have been proposed to work on LiDAR data \cite{zhou2020cylinder3d} \cite{cortinhal2020salsanext} \cite{wu2018squeezesegv2} \cite{hou2022point} 
and some with multi-modality and images data like \cite{yan20222dpass}, which achieve impressive results. All rely on large scale datasets, such as SemanticKITTI \cite{behley2019semantickitti}, nuScenes \cite{caesar2020nuscenes} \cite{fong2021panoptic} or Waymo Open Dataset \cite{sun2020scalability}. Building perception datasets is resource intensive, requiring dedicated acquisition vehicles, post-process data pipeline and annotation platforms, making entry into production a high initial cost.

We propose the following contributions in this paper:
\begin{itemize}
    \item Navya3DSeg (Fig. \ref{fig:n3ds_samples}) is a high diversity autonomous vehicle dataset built for semantic segmentation task. It was constructed with an affordable LiDAR containing over 50,000 labeled and 80,000 unlabeled frames coming from over 13 countries, 39 cities and 3 continents. Navya3DSeg is, to our knowledge, the most diverse and representative of real AV operation dataset released to this date. The point clouds were batch annotated with temporal consistency to provide precise labeling over 30 labels grouped into 6 categories. The semantic label definition on point clouds is a critical step in constructing datasets. We have included the CURB label to our point cloud dataset (frequently omitted for LiDAR in AV datasets) to enable perception tasks such as lateral localization, lane fitting and road segmentation. 
    \item A novel implementation of multi-label stratified dataset split for sequential LiDAR scan data to obtain better train, validation and test subsets.
    We demonstrate an improved train-validation split on the existing SemanticKITTI dataset with an improved test set performance in \textit{mIoU} of 1.2\% over the baseline split provided.
    \item An extensive benchmark of semantic segmentation on existing state of the art models, SqueezeSegV2 \cite{wu2018squeezesegv2}, SalsaNext \cite{cortinhal2020salsanext} and Cylinder3D \cite{zhou2020cylinder3d}.
    \item A Bayesian Active Learning (AL) benchmark to perform dataset distillation on Navya3DSeg.
    \item A novel ego-pose distance based, heuristic-free sampling method for sequential datasets.
\end{itemize}

\section{Related work}

\begin{figure*}[htb]
    \centering
    \includegraphics[width=0.85\textwidth]{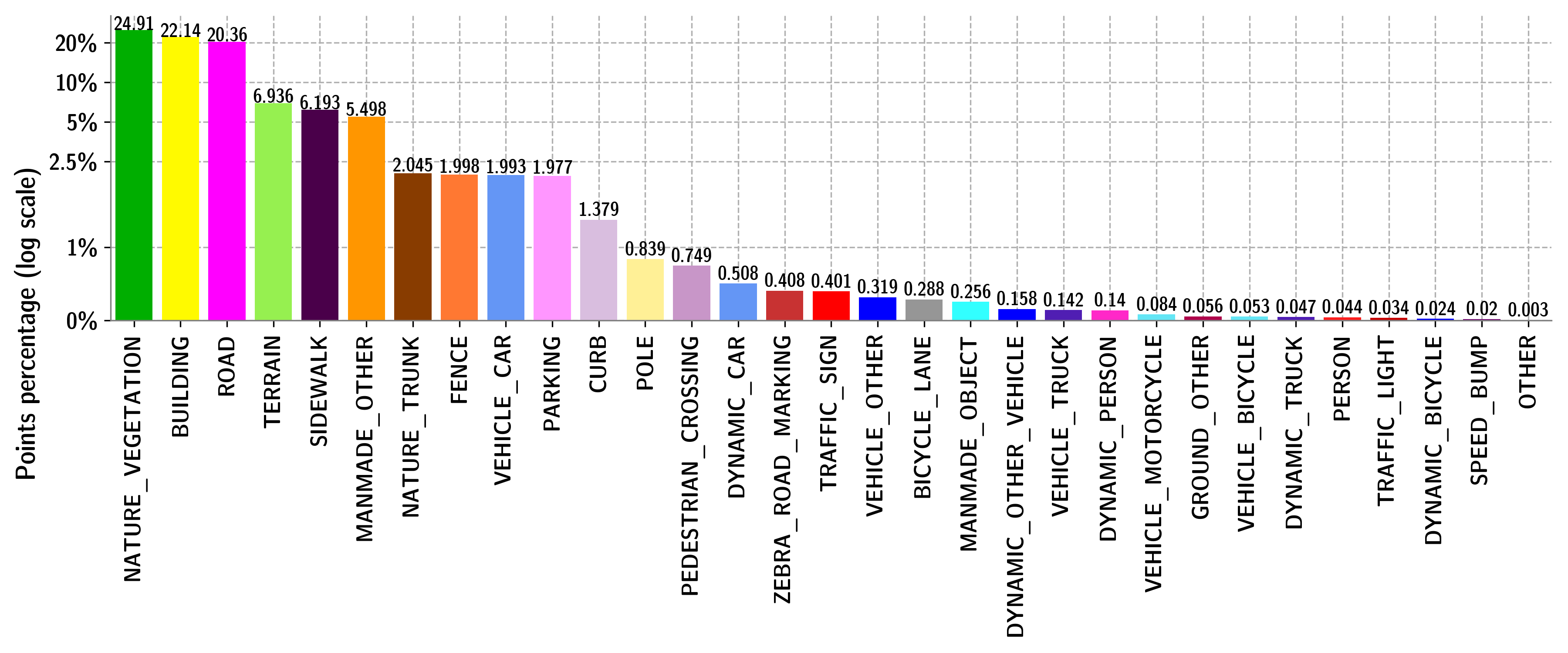}
    \caption{Label distribution of Navya3DSeg.}
    \label{fig:n3ds_labels}
\end{figure*}

Advances in LiDAR perception and point cloud related tasks have created a new wave of autonomous driving datasets. A pioneer in the field was SemanticKITTI \cite{behley2019semantickitti}, a dataset with semantic labels for 3D point clouds generated from a Velodyne HDL-64E LiDAR. Waymo releases Waymo Open Dataset (WOD) \cite{sun2020scalability}, a perception dataset with 2D and 3D semantic annotations, 2D and 3D instance bounding boxes, with sequences recorded only in the USA. Unfortunately, the sensor set is proprietary (composed of 1 long range and 4 mid range LiDARs), and the annotations on point clouds were obtained after 2D-3D re-projection from camera. Thus, 3D semantic labels are provided for a small fraction of the dataset, and only for their long range LiDAR, at a frequency of 2Hz. Thus the 3D labels are not temporally consistent and do not permit frame aggregation during model training. Motional \cite{caesar2020nuscenes} proposed the first 3D multi-modal object detection dataset, recorded in the cities of Boston and Singapore, using a Velodyne HDL-32E LiDAR, with an extension called Panoptic nuScenes \cite{fong2021panoptic}, while SemanticPOSS \cite{pan2020semanticposs} dataset focuses on dynamic labels. Finally, KITTI-360 dataset \cite{liao2022kitti} contains semantic segmentation and bounding box annotations in both image and LiDAR modalities (Velodyne HDL-64E and a SICK LMS 200 laser). A recent point cloud semantic segmentation dataset, HelixNet, using a Velodyne HDL-64E LiDAR, focusing mainly on french locations was released by \cite{loiseau22online}. Table \ref{tab:dataset_comparison} compares Navya3DSeg to the aforementioned datasets.

Point cloud datasets are key for autonomous driving perception tasks, and among them we can cite:
\begin{itemize}
    \item \textit{Semantic Occupancy Grid \& Bird Eye View (BEV) Perception}: A Semantic Occupancy Grid (SOG) is a discrete 2D-cell representation of space with an associated label. SOG and BEV representations are key to perform multi-sensor fusion and upstream usage by tracking, path planning and decision making modules \cite{ma2022vision}. 
    \item \textit{Automating HDMap creation}: Semantic point clouds are useful in automating various sub-tasks in HDMaps creation: lane, curbs, road edge and topology extraction, depth and elevation map extraction \cite{liu2020high}. Semantic point clouds also help improve global localization \cite{garg2020semantics}.
\end{itemize}


\begin{table}[h]
\caption{Dataset comparison}
\label{tab:dataset_comparison}
\resizebox{\columnwidth}{!}{
\begin{tabular}{ccccc}
\textbf{Name} &  \textbf{\# labeled scans} &  \textbf{LiDAR sensor set} & \textbf{Countries} & \textbf{Labels} \\ \hline
SemanticKITTI       & 43552 & Velodyne HDL-64E          & 1 (Germany) & 28 \\
SemanticPOSS        & 2988  & Hesai Pandora & 1 (China)   & 14 \\
KITTI-360           & 100000  & Velodyne HDL-64E & 1 (Germany)   & 19 \\
nuScenes            & 40000 & Velodyne HDL-32E           & 2 (USA, SG) & 32 \\
Waymo Open Dataset  & 46000  & 1 in-house LiDAR         & 1 (USA)     & 23 \\
Navya3DSeg                & 50070 & Hesai Pandar40            & 9           & 30
\end{tabular}
}
\end{table}

\begin{figure*}
    \centering
    \includegraphics[width=0.90\linewidth]{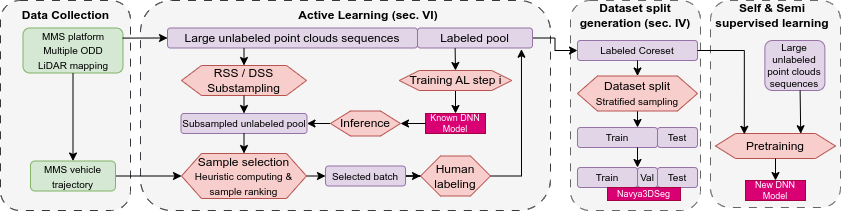}
    \caption{An overview of the data collection, AL coreset extraction with human in the loop, and dataset splitting steps.}
    \label{fig:AL-to-split}
\end{figure*}

\section{Navya3DSeg dataset}
\label{section:dataset}

Navya's autonomous shuttle operates on a varied operational domain, including rural, urban, coastal and industrial sites. The Navya3DSeg dataset was designed for autonomous shuttles operating in these special environments. It was created with an affordable but production grade mapping system, which is modular and usable with any vehicle. The Navya3DSeg dataset was built for the task of 3D semantic segmentation to automate the task of semantic 3D maps generation.

\textbf{Dataset size and diversity:} Navya3DSeg is split into a labeled and an unlabeled part. The labeled part is made of carefully selected 23 sequences from 9 countries (Japan, China, Norway, Denmark, Germany, Switzerland, France, Israel, USA), over 20 different cities, for a total of more than 1.5 billion points. The unlabeled part contains 25 sequences from 8 countries (Japan, South Korea, France, Norway, Saudi Arabia, UAE, USA, Scotland) and 20 cities, for a total amount of more than 2.5 billion points. With an accumulated path length of more than 130 kilometers, the sequences were collected across various environments that include urban, country side and residential areas, but also scenes with multi-lanes roads, roundabouts, intersections with cars, trucks, pedestrians in traffic, etc.

\textbf{Hardware and sensors setup for dataset acquisition:} Navya3DSeg point clouds were collected using Navya's in-house mapping platform, called Mobile Mapping System (MMS). It hosts an affordable single LiDAR sensor, a Hesai Pandar40 with 40 layers, a vertical and horizontal field of view of 23° (-16° to 7°) and 360°, and a vertical and horizontal angular resolution of 0.33° and 0.4°. On MMS, the LiDAR has an orientation of 15° towards the ground. This pitch provides a focus on the ground and reduces samples of dynamic objects in front of the ego vehicle, which is essential to build maps for localization purpose. An iXblue ATLANS-C Inertial Navigation System (INS) provides high precision position and orientation data, leading to registered point clouds without any explicit SLAM or registration algorithms. 

\textbf{Labels:} The human expert based annotation was conducted jointly with Playment, Telus International. The annotation pipelines for autonomous perception were well established, with several verification steps to evaluate the quality of annotations. We associate 25 static and 5 dynamic labels to each point in the Navya3DSeg dataset. These 30 labels are grouped into 6 categories: Ground, Vehicle, Manmade, Human, Nature and Dynamic. Navya3DSeg contains (see Fig. \ref{fig:n3ds_labels}) a large amount of label imbalance due to the natural distribution between objects and features observed in the real world, notably, the frequency of 20 different labels is less than 1\% of the dataset. Navya3DSeg dataset introduces the CURB label which is missing in most large scale semantic segmentation datasets. Bei et al. \cite{bai2022build} highlight the lack of a dedicated CURB label in point cloud datasets, and discuss classical processing methods to extract them at a large scale. The CURB label represents 1.379\% of Navya3DSeg, namely 21,45 million points coming from various environments, from national road to urban areas, including some complex cases like crossroad and driveways which are useful for HDMap and road map generation.

\textbf{Dataset design and benchmarking: } Dataset extension with AL \cite{haussmann2020scalable} has been important in industrial research, where the goal of a dataset is to provide a diversified and comprehensive large-scale ground truth for point cloud semantic segmentation task. We focus on equally important problems of dataset extension and coreset selection with AL. This enables us to extend existing datasets over time while reducing redundancy and improve model performance. We demonstrate ego-vehicle trajectory based sampling and model uncertainty driven data collection policies. Finally at the end of this pipeline we apply dataset splitting to create train, validation and test subsets. The Navya3DSeg's unlabeled pool is provided for baseline pre-training experiments on semi and self-supervised tasks, that could enable improvement over the primary semantic segmentation task performance on the original labeled coreset, see Fig. \ref{fig:AL-to-split}.

\section{Dataset split generation}
\label{section:split}
Splitting datasets is a key hidden engineering step for dataset construction through the direct impact on model's capacity to generalize. Dataset splitting partitions a large dataset into train, validation and test subsets. The split should ensure a training subset with large diversity to fit large models, a validation subset to fit hyper-parameters for model selection, and finally a test subset to evaluate the generalization performance of the optimal (on validation subset) model. Concretely, stratification is employed for dataset splits. Stratified splits are defined as k-way partitions of full dataset such that the joint distribution of the labels is conserved, while respecting the subset target ratios. Most stratified algorithms preserve the unitary label frequency, especially to avoid rare joint label redistribution cases. Sechidis et al. \cite{sechidis2011stratification} propose Multi-label Stratified Shuffle Split (MSSS), where each sample is associated with multiple labels, operating on label-indicator function per sample, without their numerical frequency. Higher order interactions between labels were studied in \cite{szymanski2017network}.
Yogamani et al. \cite{yogamani2019woodscape} presented the Woodscape dataset and were one of the firsts to address this problem on fish-eye perception dataset, and applied a stratified dataset split strategy. They also provide multiple splits to achieve consistent hypothesis evaluation.

All consequent analysis, benchmarks and iterations on the model are influenced by this split. As Uricar et al. \cite{uricar2019challenges} mention, one of the key problem in using traditional random sampling based splits is the manifestation of law of small numbers leading to bias in creating the splits. One single instance of random sampling does not ensure a good label distribution in the subsets. This is comparable to get a different proportion of heads with a small number of flips of an unbiased coin. As one increases the number of flips, the frequency of getting heads or tail reaches the expected value of 0.5 gradually.
A poor choice in splits leads to significant label imbalance issue between the training and validation subsets. SemanticKITTI has 98\% of bicycle label points in validation set (sequence 08) and the rest in the training set. We demonstrate a case study in subsection \ref{subsec:granularity_eval}.

We repeat definitions from \cite{sechidis2011stratification} of metrics to evaluate the quality of generated splits.

\begin{enumerate}
    \item Label Distribution (LD) (eq. \ref{eq:ld}) evaluates the divergence between the class-distribution in a subset of the split to the class-distribution of the whole dataset, for $c$-classes. Given the dataset $D$, the $k$-subsets : $S_1,...,S_k$:
    \begin{equation}
        LD = \frac{1}{c}\sum_{i=1}^{c} (\frac{1}{k}\sum_{j=1}^{k}| \frac{|S_j^i|}{|S_j|-|S_j^i|} - \frac{|D^i|}{|D|-|D^i|}|)
        \label{eq:ld}
    \end{equation}
    where $c$ is the number of labels, $D^i$ is the number of samples containing label $i$, $S_j^i$ is the number of samples in subset $S_j$ containing label $i$. We propose here an extension of LD, inverse frequency weighted label distribution (IFWLD, eq.\ref{eq:IFWLD}), which emphasizes the impact of critical labels, the main cause of label distribution shift among split subsets.
    \begin{equation}
    IFWLD = \sum_{i=1}^{c} \frac{|D|}{D_{i}}(\frac{1}{k}\sum_{j=1}^{k}| \frac{|S_j^i|}{|S_j|-|S_j^i|} - \frac{|D^i|}{|D|-|D^i|}|)
    \label{eq:IFWLD}
    \end{equation}
  
    \item We propose an input domain sensitive metric, Intensity Drift Score (IDS). It is calculated as the average Wasserstein distance (WD) between sample distributions of intensity channel values. We used Deepchecks library \cite{https://doi.org/10.48550/arxiv.2203.08491}, which implements WD from Scipy \cite{2020SciPy-NMeth}.
    \item Examples Distribution (ED) \cite{sechidis2011stratification} measures how the target subset size $\left|S_j \right|$ deviates from the desired ratio $c_j$ over the full dataset :       
    $ED = \frac{1}{k} \sum_{j=1}^k \lvert \lvert S_j \rvert  - c_j \rvert$.
    Our desired subset ratio in our experiments train:validation:test is 0.7:0.1:0.2.
    \item Label Kullback-Leibler Divergence (KL) : We measure the average (over 3 subsets) KL-divergence between normalized K-class histogram $P$ for subsets, and $Q$ for full dataset . $KL(P||Q)= \sum_{c \in [1, K]} p(k)\log(\frac{p(k)}{q(k)})$, where $p(c), q(c)$ refers to probability of class $c$ in subset and dataset respectively.
\end{enumerate}

\subsection{Stratified Sampling for Sequences}
Stratified sampling is usually achieved by proportionate allocation of labels between subsets. This is seen as a greedy optimization problem. In a classification problem with single label's per image, one could employ stratified sampling from Scikit-learn \cite{scikit-learn}. In case of object detection and segmentation, each sample is associated with multiple labels, referred to as multi-label/label-set classification problems. Sechidis et al. \cite{sechidis2011stratification} introduce Multi-label Stratified Shuffle Split (MSSS)  as solution for the binary multi-label case stratification problem. 
Since we work on point cloud sequences, we generalize the single sample based MSSS to sequential data by considering concatenation of consecutive scans as a single sample. This helps reduce temporal correlation between subsets. We shall term the size of this grouping to be the granularity parameter.
Furthermore in our extension, termed as Multi-label Segment Stratified Shuffle Split (MSegSSS), we use the actual frequency of labels in each segment of grouped scans, where as MSSS considers just the binary presence/absence of labels. In standard MSSS implementation, Sechidis et al. \cite{sechidis2011stratification} iteratively assign samples based on label frequency (going from least to most frequent) to the different k-splits. While in our method, due to the effect of aggregation of samples into larger segments, several labels land-up with similar frequencies, we choose the least present sequence associated with samples to assign to splits in such a case. 
Smaller granularity segments produces smaller input-output domain divergence between subset and dataset. When working with complete sequences of variable size, granularity is variable, an advantage of the MSegSSS algorithm. 

\begin{figure}[h]
    \centering
    \includegraphics[width=0.9\linewidth]{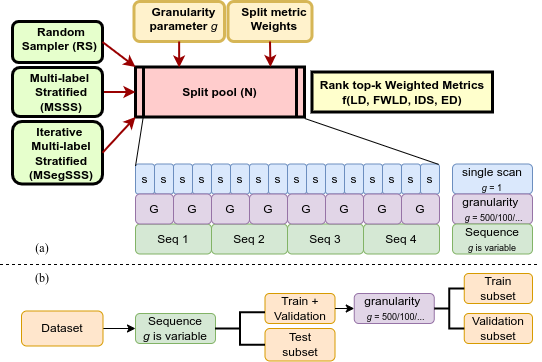}
    \caption{(a): Split module with granularity. (b): Train, Validation and Test subsets creation pipeline.
    }
    \label{fig:splitmodule}
\end{figure}

\textbf{Sampling methods:} Our split module (Fig. \ref{fig:splitmodule}) contains three sampling methods to generate a split, which are Random, MSSS and MSegSSS. The Navya3DSeg test set, representing our target domain, was built with a \textit{sequence level} granularity, meaning that each subset is composed of complete sequences. The test set will remain fixed, though different train/validation subsets can be generated at different granularity.

\textbf{Split procedure:} We use the aforementioned methods to generate a pool of multiple splits. The splits are ranked by the final objective function evaluated by the weighted combination of split metrics (LD, FWLD, IDS, ED). The selected split is the one on top of the ranking table, closest to the combination of our objectives. Navya3DSeg default split is generated by following this procedure with granularity as sequence size. 

When compared to SemanticKITTI dataset split, the selected split of Navya3DSeg generated by our split module has lower values of IDS and LD, as shown in table \ref{tab:split:n3ds_vs_sk}, thanks to the use of stratification for split creation for Navya3DSeg.

\begin{table}[h]
\centering
\captionsetup{justification=centering}
\caption{Metrics results from the dataset split module}
\begin{tabular}{c|c|c}
 Dataset      & LD (train, validation ) & IDS (train, validation, test)\\
\hline
Navya3DSeg          &  0.00512  &   0.005861 \\
SemanticKITTI &  0.00809 &    0.049727\\
\end{tabular}
\label{tab:split:n3ds_vs_sk}
\end{table}

\subsection{Sampling methods evaluation}
We performed a benchmark of sampling methods for dataset split creation of train and validation subsets, with different granularity levels. For this, we generate 10000 different splits with a desired ratio of 0.8:0.2 (train:validation) for each sampling method and granularity level selected, and evaluate them by our split metrics. From table \ref{tab:split:methods}, MSegSSS is the best method overall to sample a split satisfying the condition of label distribution. MSegSSS is slightly better than MSSS, based on LD and IFWLD. Both MSSS and MSegSSS are better than Random in all metrics. However, MSSS and MSegSSS are computationally more expensive than Random, especially MSegSSS. Thus, choosing the optimal sampling method depends on the available computation power and the objectives. In practice, MSegSSS ensures a good split with few instances of splits in the pool, while MSSS and Random could bring similar splits, as the number of instances in the pool becomes very large.


\begin{table}[h]
\captionsetup{justification=centering}
\caption{Navya3DSeg sampling method benchmark for dataset split creation for training and validation subsets}
\resizebox{\columnwidth}{!}{

\begin{tabular}{c|c|c|c|c|c}
Granularity           & Method  & mean(LD)            & mean(IFWLD)         & mean(IDS)           & mean(ED)            \\ \hline
\multirow{3}{*}{100}  & MSSS    & 2,4031E-03          & 9,5066E-06          & \textbf{2,0470E-03} & \textbf{2,3920E-02} \\
                      & MSegSSS & \textbf{2,3944E-03} & \textbf{9,5048E-06} & 2,0505E-03          & 2,3921E-02          \\
                      & RANDOM  & 2,6614E-03          & 1,0546E-05          & 2,6401E-03          & 2,6157E-02          \\ \hline
\multirow{3}{*}{500}  & MSSS    & 4,5246E-03          & 1,6637E-05          & \textbf{7,0008E-03} & 4,2733E-02          \\
                      & MSegSSS & \textbf{4,5031E-03} & \textbf{1,6629E-05} & 7,0315E-03          & \textbf{4,2695E-02} \\
                      & RANDOM  & 5,0154E-03          & 1,7803E-05          & 8,1847E-03          & 4,4778E-02          \\ \hline
\multirow{3}{*}{2000} & MSSS    & 8,2665E-03          & 2,7897E-05          & 1,6546E-02          & \textbf{7,4636E-02} \\
                      & MSegSSS & \textbf{8,2336E-03} & \textbf{2,7858E-05} & \textbf{1,6425E-02} & 7,4666E-02          \\
                      & RANDOM  & 8,4730E-03          & 2,8399E-05          & 1,7790E-02          & 7,6469E-02         
\end{tabular}

}
\label{tab:split:methods}
\end{table}

\subsection{Granularity evaluation}
\label{subsec:granularity_eval}
To evaluate the effect of granularity, we use our split module (Fig. \ref{fig:splitmodule}) to generate train/validation splits with different granularity values, then evaluate them using our split metrics (LD, IDS) and mIoU metrics on test set. We chose SalsaNext as the training architecture and the best model weights are selected based on \textit{mIoU} evaluated on validation set for each split. The table \ref{tab:split:granularity_exps} shows experiments on different train/validation splits, corresponding to different granularity, and evaluated on the same test set of each dataset.

\textbf{Navya3DSeg:} Smaller granularity results in artificially high performance scores on both train and validation subsets. The worst case is granularity of 1. This is due to the temporal correlation between scans in training and validation sets, and leads to subsequent poor test set generalization. The ideal split extraction is a constrained optimization problem, which minimizes temporal correlation between subsets, as well as minimizes split metrics for given granularity. The higher the granularity, the lower the temporal correlation between subsets, but at the same time, splits might not reach the target ratios for subset size. For both
SemanticKITTI and Navya3DSeg, the test subset is made with a split at sequence level. But for Navya3DSeg, it is defined with stratification.

In Table \ref{tab:split:granularity_exps}, we obtain two splits with the same granularity of 2000, but with different obtained subset size ratios (different from the targeted ratios, since obtained ratio achieved is output sensitive). The $2000^\dagger$  granularity split with targeted ratio aligned with sequence level split, has higher scores on validation and lower on test, with lower KL divergence, with small loss in IDS scores. While the 2000 granularity split shows that the algorithm can produce obtained split ratios not respecting the input targeted ratios, leading to poorer performances across metrics.

\textbf{SemanticKITTI} original dataset split is imbalanced with a validation set (sequence 08) being the dataset's most diverse and longest sequence. Especially, it contains 99.2\% of static bicyclists labels in the dataset. There are high intensity drift between validation and train sets, and also between validation and test sets. We take advantage of our split module to find another train/validation split with lower label divergence and intensity shift, while having similar sizes of train and validation sets. As a result, the models trained on splits with granularity of 500 and 1000 have better performance on test set compared to the one trained on the default split, with a 1.2\% \textit{mIoU} boost when granularity is set to 500.

\begin{table*}[ht]
\centering
\captionsetup{justification=centering}
\caption{Granularity evaluation on Navya3DSeg and SemanticKITTI. The sequence split (Sequence*) of SemanticKITTI was taken from the dataset configuration}
\scalebox{0.96}{
\begin{tabular}{c|cc|ccccccc}
\multirow{2}{*}{Dataset}       & \multicolumn{2}{c|}{Input variables} & \multicolumn{7}{c}{Output metrics}                                           \\ \cline{2-10} 
                               & Granularity     & Targeted ratios    & Obtained ratios    & LD $\downarrow$      & IDS $\downarrow$   & KL Divergence $\downarrow$    & Train mIoU & Val mIoU & Test mIoU \\ \hline
\multirow{5}{*}{Navya3DSeg}          & Sequence        & 87.5:12.5          & \textbf{79.8:20.2} & 5,10E-03 & 5,90E-03 & 6,45E-03 & 73.15        & 45.67    & 50.69     \\
 & 2000 & 87.5:12.5          & 87.3:12.7 & 4,60E-03 & 1,80E-02 & 7,01E-03 & 69.03 & 47.27 & 48.30 \\
 & $2000^\dagger$ & \textbf{79.8:20.2} & 79.5:20.5 & 4,30E-03 & 6,00E-03 & 4,27E-03 & 72.05 & 47.93 & 49.37 \\
 & 500  & \textbf{79.8:20.2} & 79.0:21.0 & 4,70E-03 & 3,60E-03 & 1,62E-03 & 72.77 & 54.59 & 50.49 \\
 & 1    & \textbf{79.8:20.2} & 79.7:20.3 & \textbf{2,00E-04} & \textbf{1,00E-04} & \textbf{7,00E-06} & 72.10 & 72.72 & \textbf{51.09} \\ \hline
\multirow{3}{*}{SemanticKITTI} & Sequence*         & -                  & \textbf{82.7:17.3} & 8,10E-03 & 6,53E-02 & 1,87E-02 & 80.84        & 53.66    & 54.81     \\
 & 1000 & \textbf{82.7:17.3} & 82.0:18.0 & 6,80E-03 & 3,50E-03 & 1,31E-02 & 81.62 & 61.43 & \textbf{56.00} \\
 & 500  & \textbf{82.7:17.3} & 82.5:17.5 & \textbf{3,40E-03} & \textbf{2,70E-03} & \textbf{4,35E-03} & 82.47 & 64.03 & 55.97
\end{tabular}
}
\label{tab:split:granularity_exps}
\end{table*}

\section{Semantic segmentation benchmark}
\label{section:benchmark}
\subsection{Navya3DSeg semantic segmentation benchmark}
Several labels have a very low frequency in Navya3DSeg (Fig. \ref{fig:n3ds_labels}); this is why we employ a label mapping to group certain labels into one single label to get a more balanced label distribution over the dataset. As an example, the label BICYCLE\_LANE is mapped to ROAD. This label mapping is used in experiments at sections \ref{section:split}, \ref{section:benchmark} and \ref{section:al_benchmark}. We present here the benchmark results on the semantic segmentation task, using range image based representation network architectures such as SalsaNext \cite{cortinhal2020salsanext}, SqueezeSegV2 \cite{wu2018squeezesegv2}, and 3D cylindrical based representation network Cylinder3D \cite{zhou2020cylinder3d}, which takes advantage of the full 3D sparsity of the point clouds. For range image based representation networks, we apply a spherical projection \cite{wu2018squeezesegv2} on point clouds to obtain 2D range images with a resolution of 896x40 (SqueezeSegV2) and 900x40 (SalsaNext) as inputs for the network. With Cylinder3D, we use the default cylindrical grid configuration of size [480, 360, 32] to build the cylindrical partition which keeps an uniform point distribution within voxels across distance. Cylinder3D is the most effective architecture, as seen on table \ref{tab:bench_class_iou}. It is noted that Navya3DSeg default split (section \ref{section:split}) is used for all experiments in sections \ref{section:benchmark} and \ref{section:al_benchmark}.

In order to evaluate Navya3DSeg on different architectures, we use standard metrics for semantic segmentation tasks, namely \textit{Intersection over Union (IoU)}, \textit{Precision-Recall} and \textit{F1-Scores}. Following Fong et al. \cite{fong2021panoptic}, for each metric we also compute the mean and frequency weighted average over every label for each set (validation, test) and subset (\textit{things}, \textit{stuff}).

\definecolor{amaranth}{rgb}{0.9, 0.17, 0.31}
\definecolor{blush}{rgb}{0.87, 0.36, 0.51}
\definecolor{darkolivegreen}{rgb}{0.33, 0.42, 0.18}
\definecolor{darktangerine}{rgb}{1.0, 0.66, 0.07}
\definecolor{darktan}{rgb}{0.57, 0.51, 0.32}

\def\cmiou{\fcolorbox{black}{black}{\rule{0pt}{3pt}\rule{3pt}{0pt}}\:}
\def\ccar{\fcolorbox{NavyBlue}{NavyBlue}{\rule{0pt}{3pt}\rule{3pt}{0pt}}\:}
\def\cbicycle{\fcolorbox{SkyBlue}{SkyBlue}{\rule{0pt}{3pt}\rule{3pt}{0pt}}\:}
\def\cmotorcycle{\fcolorbox{Blue}{Blue}{\rule{0pt}{3pt}\rule{3pt}{0pt}}\:}
\def\ctruck{\fcolorbox{RoyalPurple}{RoyalPurple}{\rule{0pt}{3pt}\rule{3pt}{0pt}}\:}
\def\cotherveh{\fcolorbox{blue}{blue}{\rule{0pt}{3pt}\rule{3pt}{0pt}}\:}
\def\cperson{\fcolorbox{red}{red}{\rule{0pt}{3pt}\rule{3pt}{0pt}}\:}
\def\cbicyclist{\fcolorbox{Magenta}{Magenta}{\rule{0pt}{3pt}\rule{3pt}{0pt}}\:}
\def\cmotorcyclist{\fcolorbox{RedViolet}{RedViolet}{\rule{0pt}{3pt}\rule{3pt}{0pt}}\:}
\def\croad{\fcolorbox{Rhodamine}{Rhodamine}{\rule{0pt}{3pt}\rule{3pt}{0pt}}\:}
\def\cparking{\fcolorbox{Lavender}{Lavender}{\rule{0pt}{3pt}\rule{3pt}{0pt}}\:}
\def\csidewalk{\fcolorbox{Fuchsia}{Fuchsia}{\rule{0pt}{3pt}\rule{3pt}{0pt}}\:}
\def\cothergr{\fcolorbox{BrickRed}{BrickRed}{\rule{0pt}{3pt}\rule{3pt}{0pt}}\:}
\def\cbuilding{\fcolorbox{Dandelion}{Dandelion}{\rule{0pt}{3pt}\rule{3pt}{0pt}}\:}
\def\cfence{\fcolorbox{darktan}{darktan}{\rule{0pt}{3pt}\rule{3pt}{0pt}}\:}
\def\cvegetation{\fcolorbox{OliveGreen}{OliveGreen}{\rule{0pt}{3pt}\rule{3pt}{0pt}}\:}
\def\ctrunk{\fcolorbox{Brown}{Brown}{\rule{0pt}{3pt}\rule{3pt}{0pt}}\:}
\def\cterrain{\fcolorbox{LimeGreen}{LimeGreen}{\rule{0pt}{3pt}\rule{3pt}{0pt}}\:}
\def\cpole{\fcolorbox{yellow}{yellow}{\rule{0pt}{3pt}\rule{3pt}{0pt}}\:}
\def\csign{\fcolorbox{Red}{Red}{\rule{0pt}{3pt}\rule{3pt}{0pt}}\:}
\def\clight{\fcolorbox{amaranth}{amaranth}{\rule{0pt}{3pt}\rule{3pt}{0pt}}\:}
\def\ccurb{\fcolorbox{blush}{blush}{\rule{0pt}{3pt}\rule{3pt}{0pt}}\:}
\def\cobject{\fcolorbox{BurntOrange}{BurntOrange}{\rule{0pt}{3pt}\rule{3pt}{0pt}}\:}
\def\cmanmadeother{\fcolorbox{darktangerine}{darktangerine}{\rule{0pt}{3pt}\rule{3pt}{0pt}}\:}

\begin{table*}[h]
\centering
\captionsetup{justification=centering}
\caption{label IoU: baseline models evaluated on test set}
\resizebox{\textwidth}{!}{\begin{tabular}{c|c|c|cccccccccccccccccccc}
 &
  \multicolumn{1}{c|}{\begin{sideways}\cmiou mIoU\end{sideways}} &
  \multicolumn{1}{c|}{\begin{sideways}\cmiou fwIoU\end{sideways}} &
  \multicolumn{1}{c}{\begin{sideways}\cbicycle BICYCLE\end{sideways}} &
  \multicolumn{1}{c}{\begin{sideways}\cbuilding BUILDING\end{sideways}} &
  \multicolumn{1}{c}{\begin{sideways}\ccar CAR\end{sideways}} &
  \multicolumn{1}{c}{\begin{sideways}\ccurb CURB\end{sideways}} &
  \multicolumn{1}{c}{\begin{sideways}\cfence FENCE\end{sideways}} &
  \multicolumn{1}{c}{\begin{sideways}\cobject MANMADE\_OBJECT\end{sideways}} &
  \multicolumn{1}{c}{\begin{sideways}\cmanmadeother MANMADE\_OTHER\end{sideways}} &
  \multicolumn{1}{c}{\begin{sideways}\cmotorcycle MOTORCYCLE\end{sideways}} &
  \multicolumn{1}{c}{\begin{sideways}\cparking PARKING\end{sideways}} &
  \multicolumn{1}{c}{\begin{sideways}\cperson PERSON\end{sideways}} &
  \multicolumn{1}{c}{\begin{sideways}\cpole POLE\end{sideways}} &
  \multicolumn{1}{c}{\begin{sideways}\croad ROAD\end{sideways}} &
  \multicolumn{1}{c}{\begin{sideways}\csidewalk SIDEWALK\end{sideways}} &
  \multicolumn{1}{c}{\begin{sideways}\cterrain TERRAIN\end{sideways}} &
  \multicolumn{1}{c}{\begin{sideways}\clight TRAFFIC\_LIGHT\end{sideways}} &
  \multicolumn{1}{c}{\begin{sideways}\csign TRAFFIC\_SIGN\end{sideways}} &
  \multicolumn{1}{c}{\begin{sideways}\ctruck TRUCK\end{sideways}} &
  \multicolumn{1}{c}{\begin{sideways}\ctrunk TRUNK\end{sideways}} &
  \multicolumn{1}{c}{\begin{sideways}\cvegetation VEGETATION\end{sideways}} &
  \multicolumn{1}{c}{\begin{sideways}\cotherveh VEHICLE\_OTHER\end{sideways}} \\ \hline
SqueezeSegV2 &
  41.46 & 
  60.89 & 
  37.07 &
  81.23 &
  70.65 &
  43.22 &
  22.3 &
  7.77 &
  31.77 &
  58.36 &
  12.08 &
  30.83 &
  27.44 &
  85.79 &
  57.52 &
  54.3 &
  8.92 &
  34.1 &
  3.11 &
  26.09 &
  69.07 &
  26.24 \\
SalsaNext &
53.35 & 
70.88 & 
48.08 &
84.7 &
76.27 &
51.24 &
35.4 &
23.74 &
42.75 &
71.43 &
19.53 &
57.34 &
46.96 &
88.04 &
65.55 &
64.83 &
28.15 &
43.46 &
6.92 &
40.47 &
77.84 &
41.04 \\
Cylinder3D &
  \textbf{62.84} & 
  \textbf{81.16} & 
  \textbf{67.39} &
  \textbf{92.26} &
  \textbf{89.51} &
  \textbf{53.60} &
  \textbf{55.24} &
  \textbf{36.01} &
  \textbf{50.35} &
  \textbf{88.32} &
  \textbf{25.96} &
  \textbf{79.12} &
  \textbf{63.36} &
  \textbf{89.62} &
  \textbf{69.74} &
  \textbf{68.27} &
  \textbf{49.66} &
  \textbf{55.10} &
  \textbf{36.84} &
  \textbf{53.08} &
  \textbf{84.14} &
  \textbf{49.21}
\end{tabular}}
\label{tab:bench_class_iou}
\end{table*}


\subsection{Cross-dataset experiments }

Here we evaluate fine-tuning between SemanticKITTI and Navya3DSeg. 
The experiment results are reported on validation set after 3D point cloud reprojection with SalsaNext architecture. There is a significant domain gap between SemanticKITTI and Navya3DSeg, mainly due to:

\begin{itemize}
  \item \textit{Sensor intrinsics}: Navya3DSeg has been acquired from a Hesai Pandar40 LiDAR while SemanticKITTI team used a Velodyne HDL-64E. These two sensors have a different number of layers, vertical Field of View (vFoV), angular resolution and intensity distribution. It results in different range-image representations.
  \item \textit{Sensor extrinsics}: Navya3DSeg LiDAR sensor is oriented with a pitch of 15° towards the ground while that of SemanticKITTI is laid flat at 0°. Objects in front and back of the vehicles sample different classes (buldings, roads, vegetation) differently, leading to a domain shift.
  \item \textit{Annotation policy}: There are common labels between both datasets, nevertheless, the annotation policy for each label may differ from a dataset to another. 
  \item \textit{Acquisition conditions}: Most of Navya3DSeg acquisitions were done by night, thus dynamic labels spatial distribution can differ from SemanticKITTI, like with the CAR label for example.
\end{itemize}

\textbf{Fine-tuning:}
We show the influence of pre-training by cross dataset fine-tuning between SemanticKITTI and Navya3DSeg. Models were fine-tuned using the entirety of source and target label domains, without freezing any layer, to learn target domain specificity. Results from table \ref{tab:finetuning} demonstrate that even with different sensors, density, orientation and layer distribution, pre-training leads to a better generalization on target domain. Pre-training also shows a faster model convergence: fine-tuning from Navya3DSeg to SemanticKITTI takes only 18 epochs to reach the best results on SemanticKITTI validation set, contrary to 74 epochs when training from scratch. Similarly, fine-tuning from SemanticKITTI to Navya3DSeg takes 96 epochs instead of 144 when training from scratch.

\begin{table}[h]
\captionsetup{justification=centering}
\caption{Cross dataset generalization results for SalsaNext. mPrec and mRec terms stand for mean Precision and mean Recall metrics respectively}
\resizebox{\columnwidth}{!}{
\begin{tabular}{c|c|cccccc}
              &  & \multicolumn{6}{c}{Target dataset}                                                                                                                    \\ \cline{3-8} 
              &  & \multicolumn{3}{c|}{SemanticKITTI}                                                   & \multicolumn{3}{c}{Navya3DSeg}                                        \\ \cline{3-8} 
\multirow{-3}{*}{Source dataset} &
  \multirow{-3}{*}{Fine-tuning} &
  \multicolumn{1}{c|}{mIoU} &
  \multicolumn{1}{c|}{mPrec} &
  \multicolumn{1}{c|}{mRec} &
  \multicolumn{1}{c|}{mIoU} &
  \multicolumn{1}{c|}{mPrec} &
  mRec \\ \hline
SemanticKITTI &
  \checkmark &
  \multicolumn{1}{c|}{58.43} &
  \multicolumn{1}{c|}{75.27} &
  \multicolumn{1}{c|}{65.71} &
  \multicolumn{1}{c|}{\textbf{50.60}} &
  \multicolumn{1}{c|}{\textbf{68.63}} &
  \textbf{61.64} \\
Navya3DSeg &
  \checkmark &
  \multicolumn{1}{c|}{\textbf{60.51}} &
  \multicolumn{1}{c|}{\textbf{76.82}} &
  \multicolumn{1}{c|}{\textbf{67.41}} &
  \multicolumn{1}{c|}{49.90} &
  \multicolumn{1}{c|}{68.20} &
  60.75 \\ \hline
SemanticKITTI &
& 
\multicolumn{1}{c|}{58.43} &
\multicolumn{1}{c|}{75.27} &
\multicolumn{1}{c|}{65.71} &
\multicolumn{1}{c|}{18.87} & 
\multicolumn{1}{c|}{34.40} & 32.05 \\
Navya3DSeg &  
&
\multicolumn{1}{c|}{16.86} & 
\multicolumn{1}{c|}{55.21} & 
\multicolumn{1}{c|}{26.29} & 
\multicolumn{1}{c|}{49.90} &
\multicolumn{1}{c|}{68.20} &
60.75 
\end{tabular}
}
\label{tab:finetuning}
\end{table}

\section{Dataset distillation with Active Learning}
\label{section:al_benchmark}
Bayesian Active Learning \cite{atighehchian2019baal} is the problem of sequentially selecting new unlabeled batches of image/point clouds and then annotating them with human in the loop to improve the performance of model's trained iteratively on growing labeled data. Bayesian AL uses the model's predictive uncertainty to construct heuristic functions that help select unlabeled data to be selected and annotated to improve the model. If a large amount of data is already annotated, AL can be used to select an iso-performing or core subset \cite{iyer2022subset} which, when trained upon, has approximately equivalent performance as if the model were trained on the full subset. In our previous work \cite{visapp22}, we show that the BALD heuristic \cite{houlsby2011bayesian} is giving the best dataset compression ratio (over SemanticKITTI dataset), and was well adapted to semantic segmentation task for point clouds.
Following the same process, we evaluate Bayesian AL methods over Navya3DSeg dataset to obtain a core subset. We also propose an alternative to the BALD heuristic, with our novel ego-pose distance based sampling method which is considering the sequential nature of data.

\subsection{Ego-vehicle trajectory regularity as heuristic function}
Heuristic functions help select the most informative samples for the model, by ranking samples from the unlabeled pool at each AL step. Random sampling is computationally cheaper since it does not look at the model's output. We make the observation that sequential datasets like SemanticKITTI and Navya3DSeg contain redundant viewpoints and thus point clouds, due to limited variation in ego-vehicle movement. To address these problems, we propose a novel \textit{ego-pose distance based} sampling method.

\textbf{Motivation}: To select frames for AL, Vondrick and Ramanan \cite{vondrick2011video} use temporal consistency measures, based on object-tracking. We propose an ego-pose (distance between ego-poses of the mapping vehicle) distance based sampling that follows a similar idea. Alike random sampling, we do not require model predictions, which is computationally costly. Though, we look at distances between ego-poses over the trajectory of the mapping vehicle to accept or reject a point cloud during the sampling step of AL. This approach assumes that smaller distance between poses produces more redundant point clouds.

\begin{figure}[!htbp]
\centering
\includegraphics[width=0.4\textwidth]{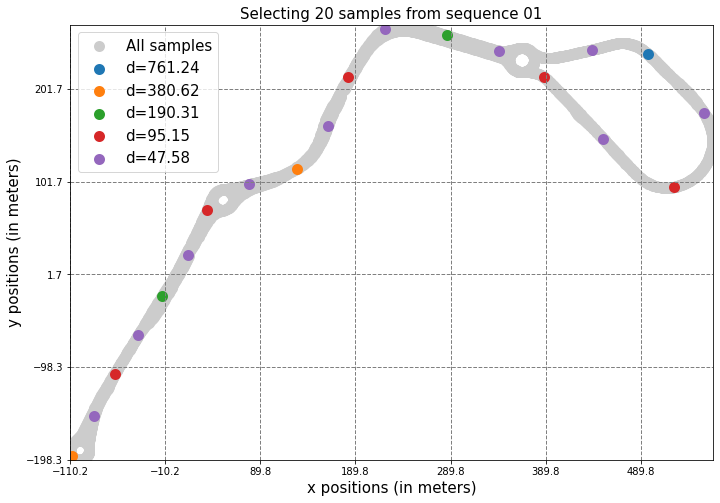}
\caption{Selected poses using ego-pose distance based sampling on Navya3DSeg sequence 01 across various distance thresholds.}
\label{fig:al:distance_sampling}
\end{figure}

\textbf{Ego-pose distance based sampling}: Our method only looks at the ego-pose information and not the point cloud itself. It selects poses that are as far away from each other, to avoid spatial correlation. Samples are chosen as a function of increasing circular neighborhoods centered on ego-poses from frames of the labeled and unlabeled pools. The neighborhood's radius sequentially spans the minimum and maximum distances between pairs of ego-poses, until every ego-pose is covered. Subsets extracted using this sampling are defined as $A_i = \{x | x \in U,  \forall y \in L,  \text{dist}(x, y) > d_i \}$, where $d_i \in [d_\text{min}, d_\text{max}]$ is a maximal distance threshold between ego-poses separating selected frames, and $A_i$ is the new query set that is added to the labeled pool for the next AL step. The distance threshold $d_i$ varies from maximum to minimum so that subsets $A_i$ are hierarchically nested subsets of poses in the Navya3DSeg dataset, as shown on Fig. \ref{fig:al:distance_sampling}.

\subsection{Active learning dataset distillation benchmark}
To evaluate the performance of AL on Navya3DSeg, we use the standard \textit{mIoU} metric for semantic segmentation, computed at the end of each AL step, on the entire label set, but also on \textit{things} and \textit{stuff} grouped labels. Besides, we measure the Labeling Efficiency (LE) \cite{beck2021effective} to compare the amount of data needed to achieve baseline performance for the different sampling techniques benchmarked, with respect to the random sampling technique.

The table \ref{tab:al:runs} presents the experiments settings. We use three different heuristics: Random Sampling (RS), our ego-pose Distance Sampling (DS) and BALD \cite{BALD2017}. To speed up this process of heuristic computation, we can apply a random subset sampling over the unlabeled pool, termed as RSS. This reduces the number of model outputs being evaluated. In the same way, the ego-pose distance is used to select a subset from large unlabeled pool, and is termed as DSS. 
Once a prior subsampling is done on the unlabeled pool, we evaluate the heuristic function on selected subset.
When using the BALD heuristic, we use the Inverse Label Frequency (ILF) to weight the heuristic score of each pixel. This reduces the impact of high frequency labels like ROAD or VEGETATION. Initial model of run (B) is the model trained on initial set sampled by run Random (A). Initial model of runs (D) and (E) is the model trained on initial set sampled by distance, which is run (C).

\begin{figure*}[]
    \centering
    \includegraphics[width=\textwidth]{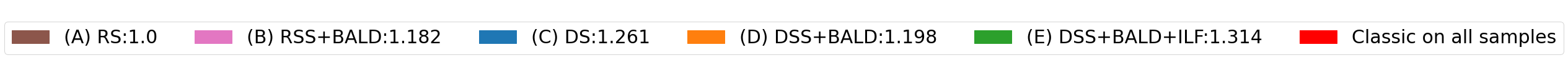}
    \includegraphics[height=3.45cm,width=0.33\textwidth]{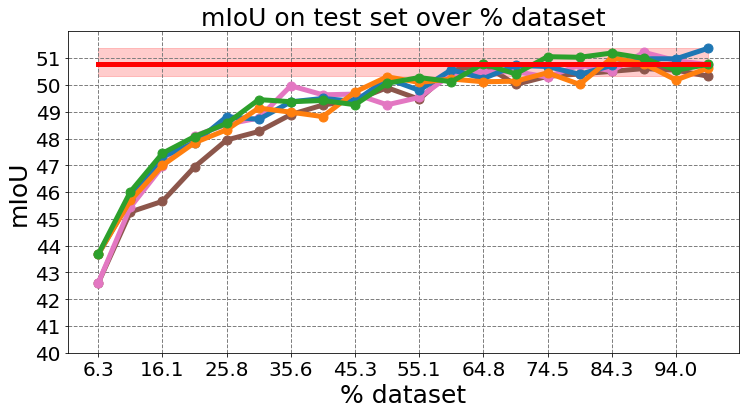} \hfill
    \includegraphics[height=3.45cm,width=0.33\textwidth]{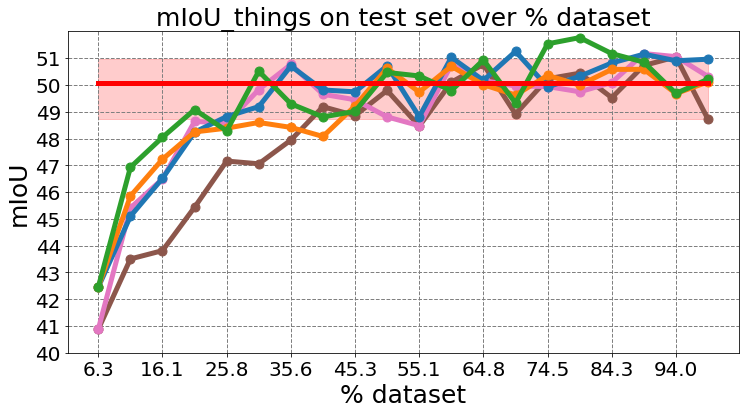}\hfill
    \includegraphics[height=3.45cm,width=0.33\textwidth]{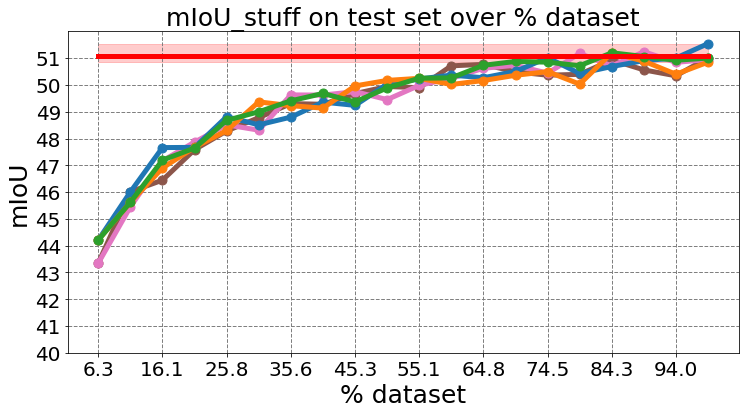}\hfill
    \caption{Navya3DSeg distillation results. Values on the legend are averages of LE ratios based on mIoU. 
    }
    \label{fig:al:results}
\end{figure*}



\begin{table}[h]
\captionsetup{justification=centering}
\caption{Active learning benchmark}
\centering
\begin{tabular}{c|c|c|c|c}
Run ID & Heuristic & \vtop{\hbox{\strut Unlabeled pool}\hbox{\strut    Sub-sampler}}  & \vtop{\hbox{\strut Label}\hbox{\strut Efficiency}}   & \vtop{\hbox{\strut Initial}\hbox{\strut model}}               \\ \hline
(A) & RS       & -   & 1.0         & \multirow{2}{*}{run (A)} \\ \cline{1-4}
(B) & BALD     & RSS & 1.182      &                          \\ \hline
(C) & DS       & -   & 1.261        & \multirow{3}{*}{run (C)} \\ \cline{1-4}
(D) & BALD     & DSS & 1.198       &                          \\ \cline{1-4}
(E) & BALD+ILF & DSS & 1.314   &                         
\end{tabular}
\label{tab:al:runs}
\end{table}

\textbf{Experiment settings:}
Following our previous work \cite{visapp22}, we use a Bayesian AL loop employing MC Dropout for uncertainty estimation. However, here we filter out pixels in range image which do not belong to any point in point clouds, then we use \emph{mean} as an aggregation function to combine all pixel-wise scores of an image into a single score. The runs with subset sampling (B), (D) and (E), sample unlabeled pool into a subset with a size of 10000 samples before computing the uncertainty scores. For each AL step, all runs (tab. \ref{tab:al:runs}) are evaluated on a fixed test set from Navya3DSeg default split (section \ref{section:split}). The rest is used for the AL pool. We use SalsaNext as the semantic segmentation architecture, with parameters from table \ref{tab:al:experiments_settings}.

\begin{table}[h]
\captionsetup{justification=centering}
\caption{Active learning experiments settings}
\begin{tabular}{cc|ccc}
\rowcolor[HTML]{EFEFEF} 
\multicolumn{2}{c|}{\cellcolor[HTML]{EFEFEF}\textbf{Cyclic Learning rate}}       & \multicolumn{3}{c}{\cellcolor[HTML]{EFEFEF}\textbf{Early stopping}}                                                                                \\ \hline
\multicolumn{1}{c|}{\cellcolor[HTML]{EFEFEF}step size}                                   & \cellcolor[HTML]{EFEFEF}policy        & \multicolumn{1}{c|}{\cellcolor[HTML]{EFEFEF}metric}                                  & \multicolumn{1}{c|}{\cellcolor[HTML]{EFEFEF}patience}                                 & \cellcolor[HTML]{EFEFEF}delta                \\ \hline
\multicolumn{1}{c|}{6 epochs}                                    & triangular1.5 & \multicolumn{1}{c|}{train  loss}                             & \multicolumn{1}{c|}{12 epochs}                                & 0.02                 \\ \Xhline{3\arrayrulewidth}
\rowcolor[HTML]{EFEFEF} 
\multicolumn{1}{c|}{\cellcolor[HTML]{EFEFEF}\textbf{batch size}} & \textbf{Initial pool} & \multicolumn{1}{c|}{\cellcolor[HTML]{EFEFEF}\textbf{budget}} & \multicolumn{1}{c|}{\cellcolor[HTML]{EFEFEF}\textbf{AL step}} & \textbf{aggregation} \\ \hline
\multicolumn{1}{c|}{34}                                          & 3055          & \multicolumn{1}{c|}{2000}                                    & \multicolumn{1}{c|}{20}                                       & sum                 
\end{tabular}
\label{tab:al:experiments_settings}
\end{table}

\textbf{Results and analysis:}
According to results in Fig. \ref{fig:al:results}, all runs reach the lower bound of baseline performance with only 60\% of data from the full dataset. Run (E) reaches the average of baseline performances with the lowest percentage of data from the full dataset, namely 64.8\%, among all runs. At the first AL steps, all runs with subset samplings - (B), (D), (E) - and ego-pose Distance Sampling (C) outperform Random Sampling (A). In general, the method from run (E) outperforms the other ones, especially on \textit{things} labels which represent only 5.06\% of the whole labeled set. We can see that the \textit{stuff} labels are more stable and have higher score than the \textit{things} labels across the different AL steps. Subsequently, new extension of the dataset should considerably help to improve models' performance on the \textit{things} labels. Regarding model initialization, models trained on the initial pool generated by our proposed ego-pose distance based method have better results than random sampling.

\section{Conclusion}
We have presented Navya3DSeg, a highly diverse large scale semantic segmentation dataset on LiDAR point clouds for production grade autonomous driving applications. We have demonstrated sequential dataset splitting using multi-label stratified sampling techniques on both Navya3DSeg and SemanticKITTI datasets, and show consequent gains in performance generalization. A benchmark on key semantic segmentation models is provided, along with a study of dataset distillation using Bayesian Active Learning framework. Our novel ego-pose distance based sampling provides a fast and efficient way to distill the dataset to 64.8\% to reach fully supervised performance on semantic segmentation task. Finally, we highlight that Navya3DSeg contains a large unlabeled pool of scans from varied operational domains of the autonomous shuttle to evaluate self-supervised and semi-supervised methods for future work.

\bibliographystyle{IEEEtran}
{\small
\bibliography{bibliography.bib}
}
%
%
%
%


%




\include{arxiv_supplementary.tex}

\end{document}

%% file: arxiv_supplementary.tex
\clearpage

\renewcommand\twocolumn[1][]{%
    \oldtwocolumn[{#1}{
    \begin{center}
		\vspace{-0.6cm}
  
      {\LARGE \bf Navya3DSeg - Navya 3D Semantic Segmentation \\ Dataset Design \& split generation for autonomous vehicles\\
- Supplementary Materials - \\ \vspace{0.1cm}
}

        
		\resizebox{.99\textwidth}{!}{%
		\includegraphics[height=2.9cm]{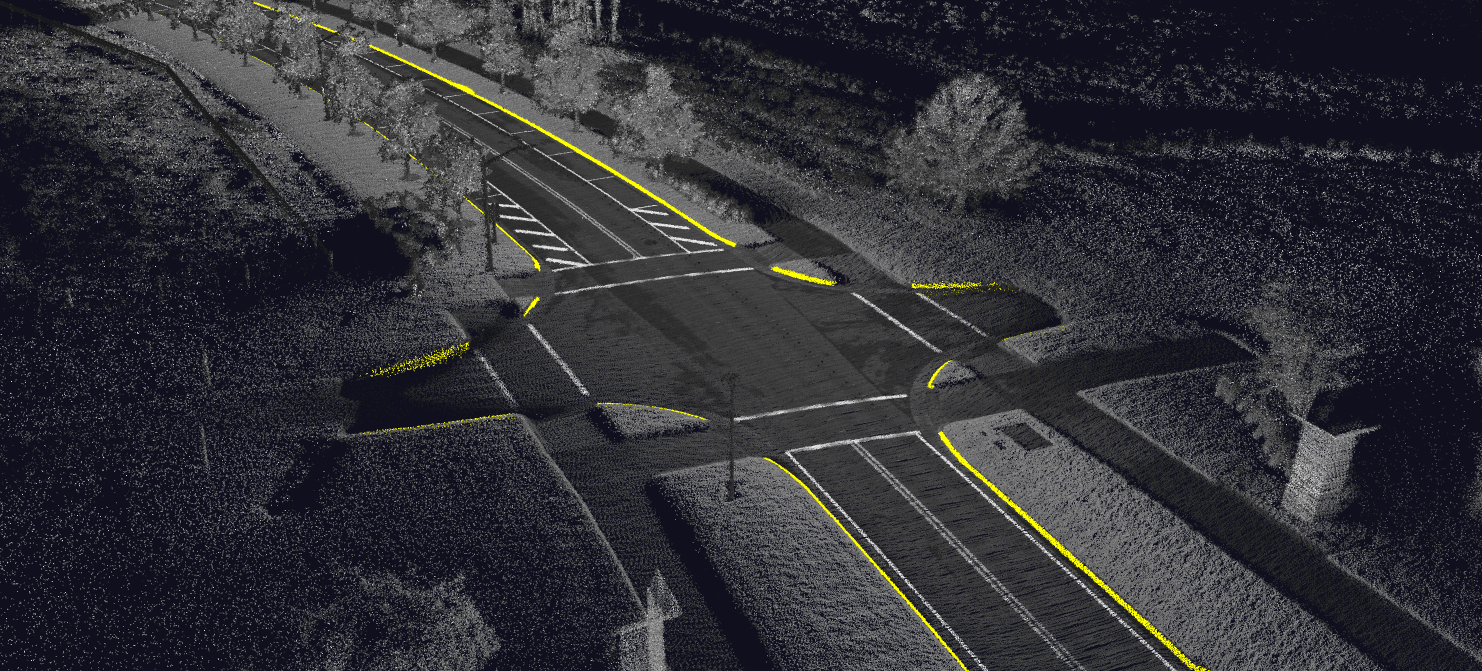} \vspace{-0.65cm} \hfill
        \includegraphics[height=2.9cm]{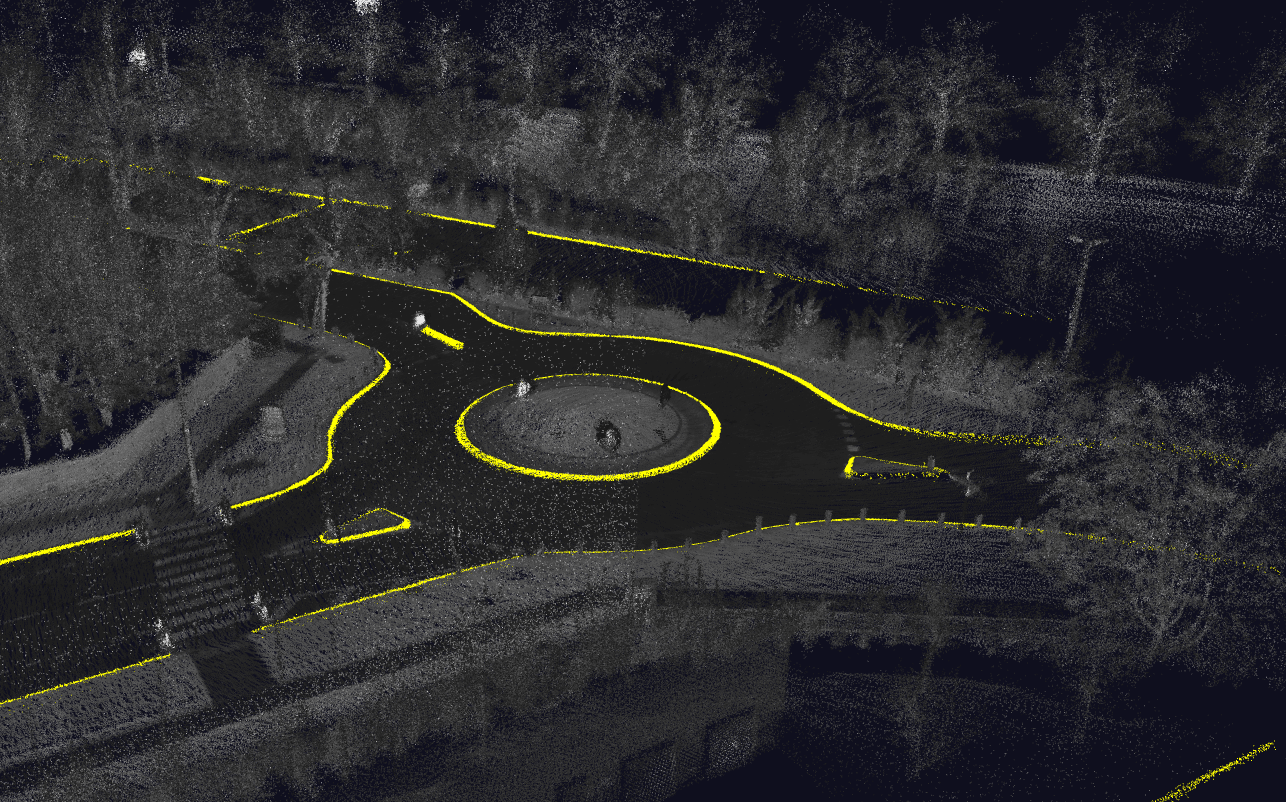} \vspace{-0.65cm} \hfill
        \includegraphics[height=2.9cm]{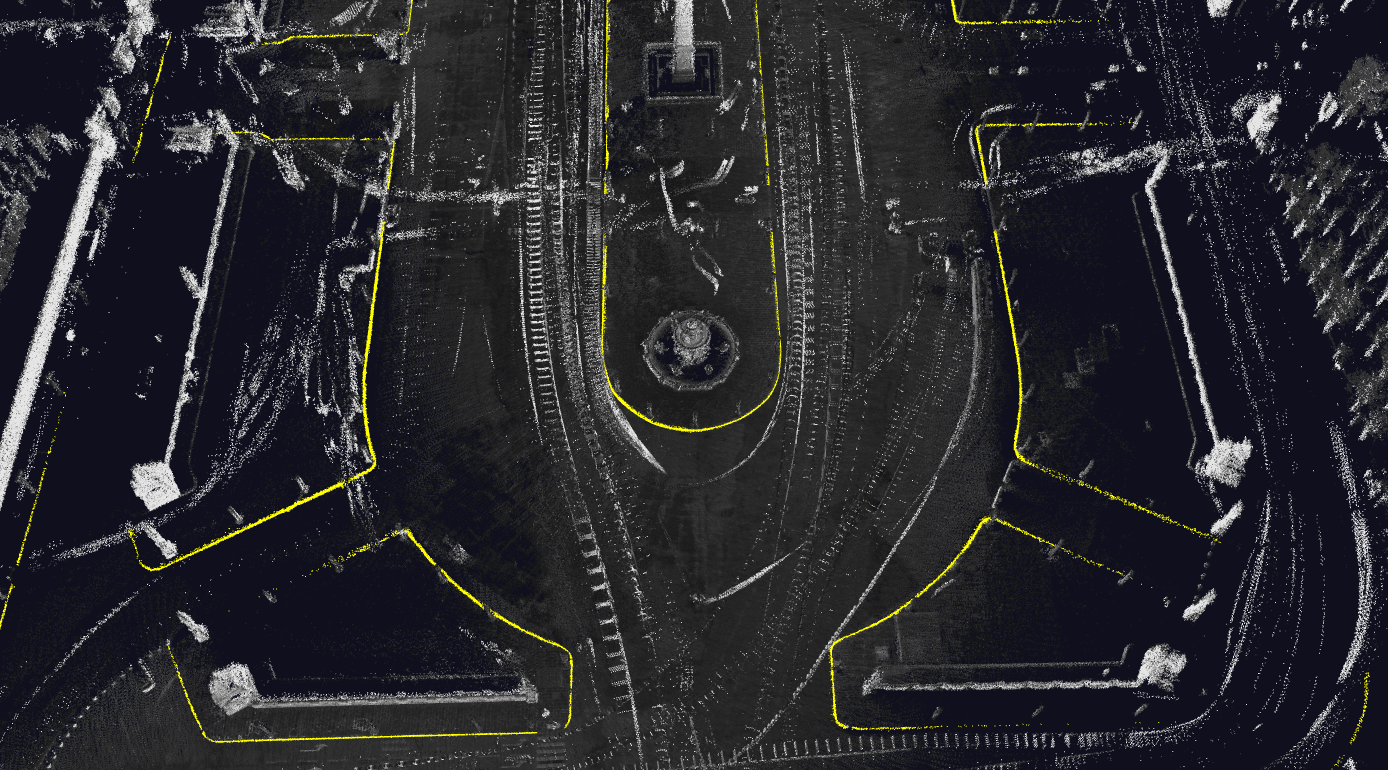}
        \vspace{-0.65cm}
        }
		\captionof{figure}
		{Navya3DSeg complex curbs examples, with the \textit{curb} label highlighted in yellow.}
		\label{fig:curbs}
	\end{center}
    }]
}

\twocolumn

\section{Navya3DSeg dataset}
The Navya3DSeg dataset provides production grade labeled point clouds for the task of semantic segmentation.
The semantic descriptions of the different labels in the dataset are key to understand the features and objects that 
are annotated. The Navya3DSeg labels' definitions, provided in Tab.\ref{tab:n3ds_Labels_definition}, were used to define the dataset's annotation policy.


Navya3DSeg dataset introduces the CURB label which is missing in most large scale semantic segmentation datasets. Bai et al. \cite{bai2022build} highlight the lack of a dedicated CURB label in point cloud datasets, and discuss classical processing methods to extract them at a large scale. The CURB label represents 1.379\% of Navya3DSeg, namely 21,45 million points coming from various environments, from national road to urban areas, including some complex cases like crossroad and driveways which are useful for HDMap and road map generation. Fig. \ref{fig:curbs} shows Navya3DSeg sequences with a highlight on the CURB label.


\section{Dataset split generation}
\subsection{Stratified Sampling for Sequences}

In standard MSSS implementation (Algo.\ref{alg:msss}), Sechidis et al. \cite{sechidis2011stratification} iteratively assign samples based on label frequency to the different k-splits, going from least to most frequent in the remaining samples. Given the subset index j and the class index i, MSSS starts by calculating the remaining desired number of samples $c_j$ and desired number of samples per label  $c^i_j$ for each subset (train, val, test). In each iteration, the algorithm examines label $l$ having the fewest number of remaining samples (line~\ref{algo:MSegSSS:label}) and stratifies its samples into output subsets $S_m$. The output subset $S_m$ for each sample is selected based on the most desired number of samples $c^i_j$ (line.~\ref{algo:MSegSSS:subset:label})  then $c_j$ (line.~\ref{algo:MSegSSS:subset}). 

Due to the effect of aggregation of samples into larger segments, several labels land-up with similar frequencies with standard MSSS implementation which considers only the binary presence/absence of label in large segment. For example, in case of a split by sequence, the number of occurrences of a label in every sequence is the number of sequences containing that label, which might end up to be the same for several labels. In this case (line.~\ref{algo:MSegSSS:label}), we choose the label having a distribution across all samples furthest away from the uniform distribution instead of the number of presences/absences of label in all segments.

\begin{algorithm}[!h]
\label{ref:iterativeStratification}
\SetKwInOut{Input}{Input}
\SetKwInOut{Output}{Output}
\SetKwComment{Comment}{/* }{ */}
\caption{IterativeStratification $(D,n, r_1 . . . r_n)$}
\label{alg:msss}
\Input{A set of instances, D, annotated with a set of labels $L = {\lambda_1, ..., \lambda_q }$,
desired number of subsets k, desired proportion of samples in each
subset $r_1,...r_k $}
\Output{Disjoint subsets $S_1, . . . S_k$ of D}

\Comment{\scriptsize Calculate the desired number of samples at each subset}
\For{$j \leftarrow 1$ \KwTo $k$}{
   $c_j = |D|r_j$\\
}

\Comment{\scriptsize Calculate the desired number of samples of each label at each subset}
\For{$i \leftarrow 1$ \KwTo $L$}{
    \For{$j \leftarrow 1$ \KwTo $k$}{
    $c^i_j \leftarrow |D^i|r_j$\\
    }
}
\While{$|D| > 0$}{
    \Comment{\scriptsize Find the label with the fewest (but at least one) remaining samples}
    $l \leftarrow \textbf{argmin}_i( |D^i|)$ \label{algo:MSegSSS:label}
    
    \ForEach{$(x, Y)\in D^L$}{
        \Comment{\scriptsize Find the subset(s) with the largest number of desired samples for this label}
        $M \leftarrow \textbf{argmax}_{j=1...k}( c^l_j)$\\ \label{algo:MSegSSS:subset:label}
        \uIf{$|M| = 1$}
        {
            $m \in M$\\
        }
        \Else{
            \Comment{\scriptsize Prioritize the subset(s) with the largest number of desired samples}
            $M' \leftarrow \textbf{argmax}_{j\in M}(c_j)$\\\label{algo:MSegSSS:subset}
            \uIf{$|M'| = 1$}
            {
                $m \in M'$\\
            }
            \Else{
                
                $m \leftarrow \textit{randomElementOf}(M')$\\
                
            }
        }
        
        $S_m \leftarrow S_m  \bigcup (x, Y) $\\
        $D \leftarrow D \backslash \{(x, Y)\}$\\
        
        \Comment{\scriptsize Update desired number of samples}
        \ForEach{$\lambda_i  \in Y$}{
           $c^i_m \leftarrow c^i_m - 1$\\    
        }
       $c_m \leftarrow c_m - 1$\\    
    }
}

\end{algorithm}

\subsection{Navya3DSeg split}

\begin{figure}[]
    \centering
    \includegraphics[width=0.49\textwidth]{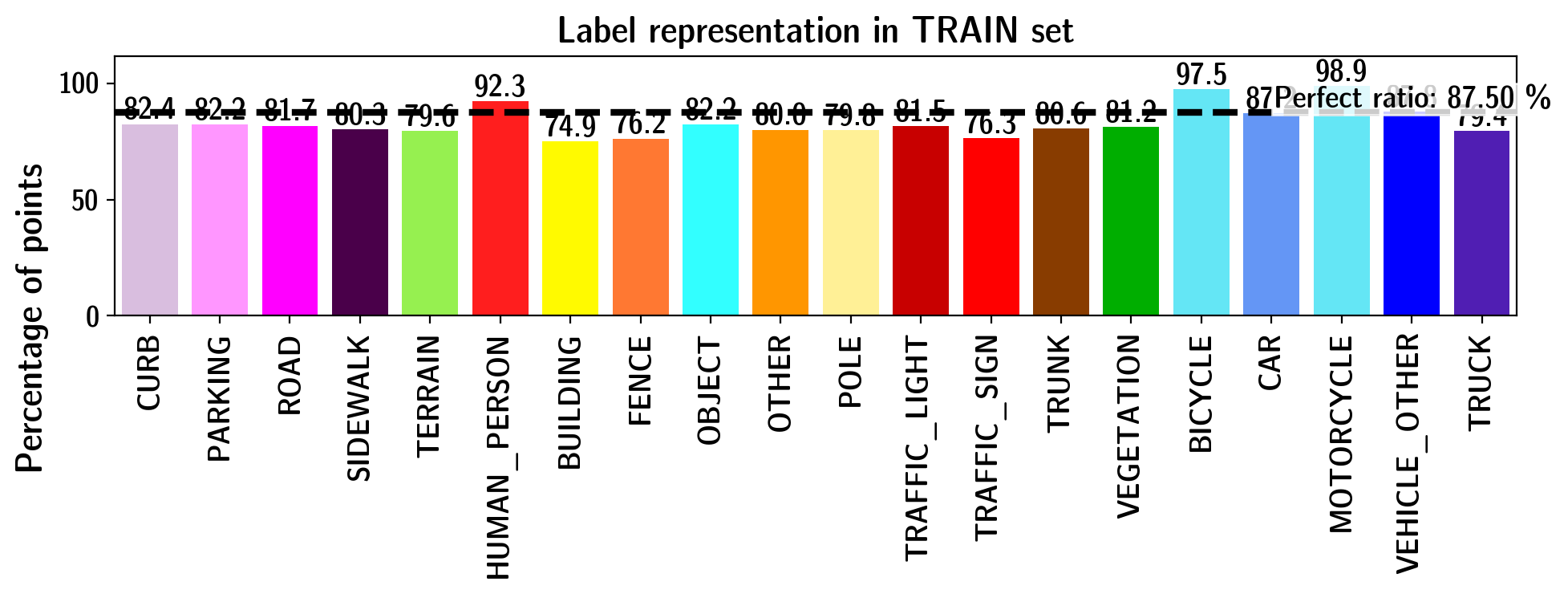}
    \includegraphics[width=0.49\textwidth]{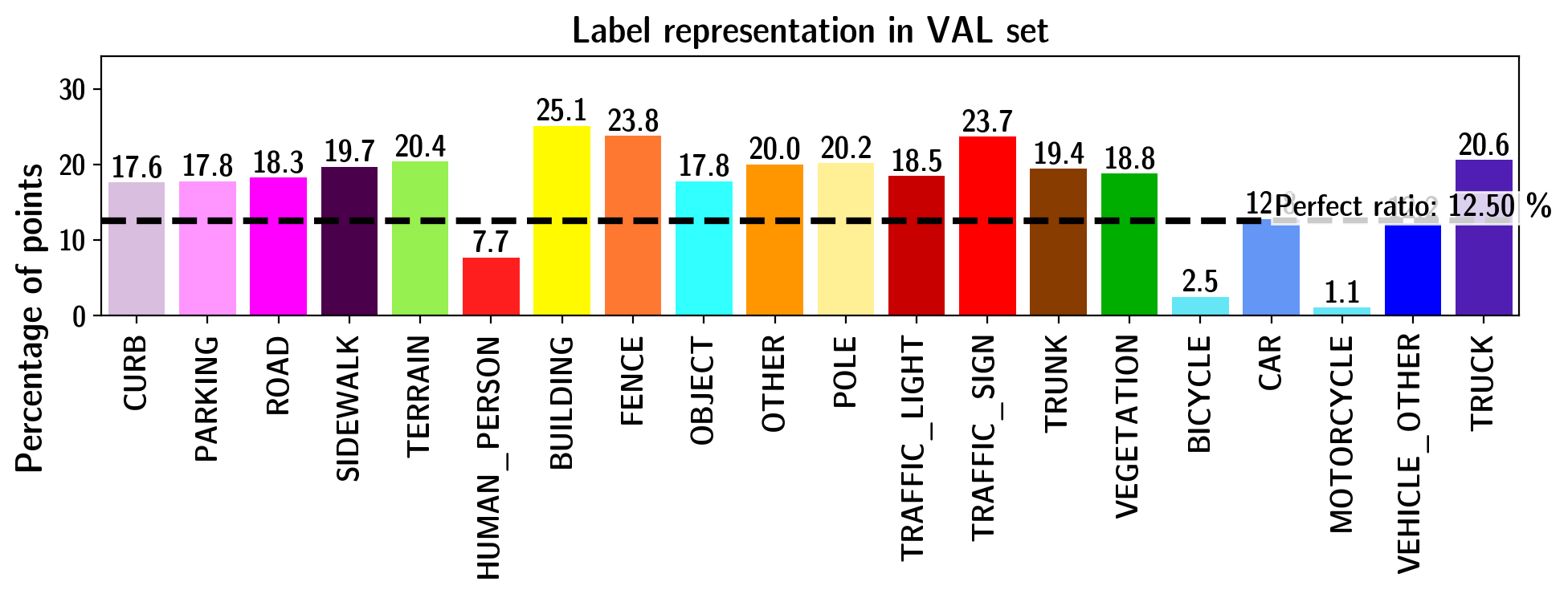}
    \captionsetup{justification=centering}
    \caption{Label representation of Navya3DSeg selected split with a desired ratio of 87.5:12.5 for train:validation subsets.}
    \label{fig:n3ds_split_seq}
\end{figure}

Following our strategy for dataset split choice, the Navya3DSeg split (Fig. 
\ref{fig:n3ds_split_seq}) is the one that satisfies the most of our multiple 
objectives. With a desired ratio of 0.7:0.1:0.2 for train:validation:test subsets 
and a sequence level granularity, we selected the sequences 12, 15, 18, 19 and 22 as 
our test set. The actual ratio of train:validation is 79.8:20.2 despite the 
87.5:12.5 desired ratio. This shift is the consequence of the sequence level 
granularity, which was mandatory to preserve a consistent test set, which labels 
will remain hidden. The figure \ref{fig:n3ds_split_seq}, namely label 
representation, shows the ratio of each label in the train and validation subsets over the 
full Navya3DSeg.

\begin{figure}[]
    \centering
    \subfloat[\centering Navya3DSeg]{{\includegraphics[width=0.49\textwidth]{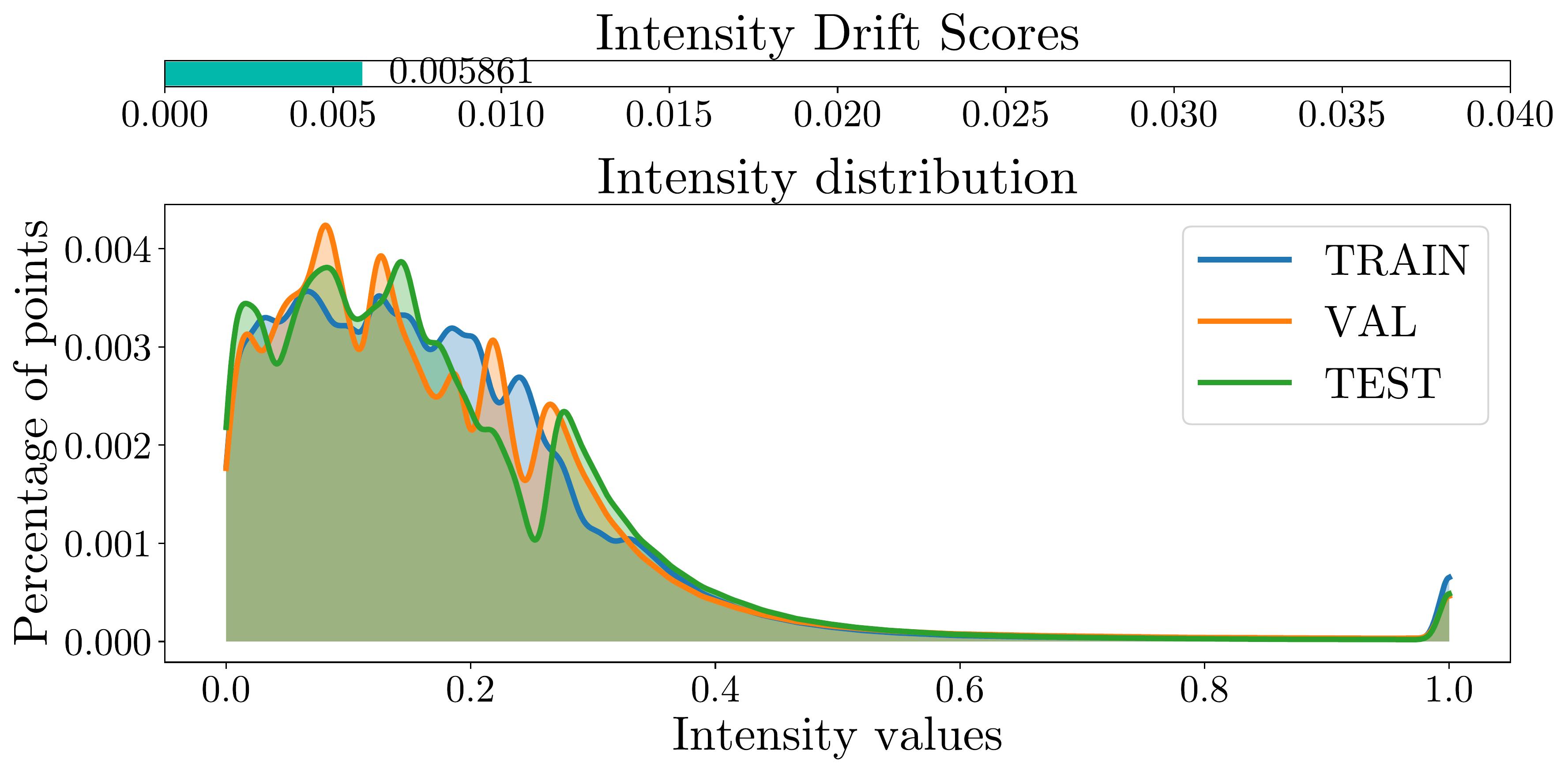}}}
    
    \subfloat[\centering SemanticKITTI]{{\includegraphics[width=0.49\textwidth]{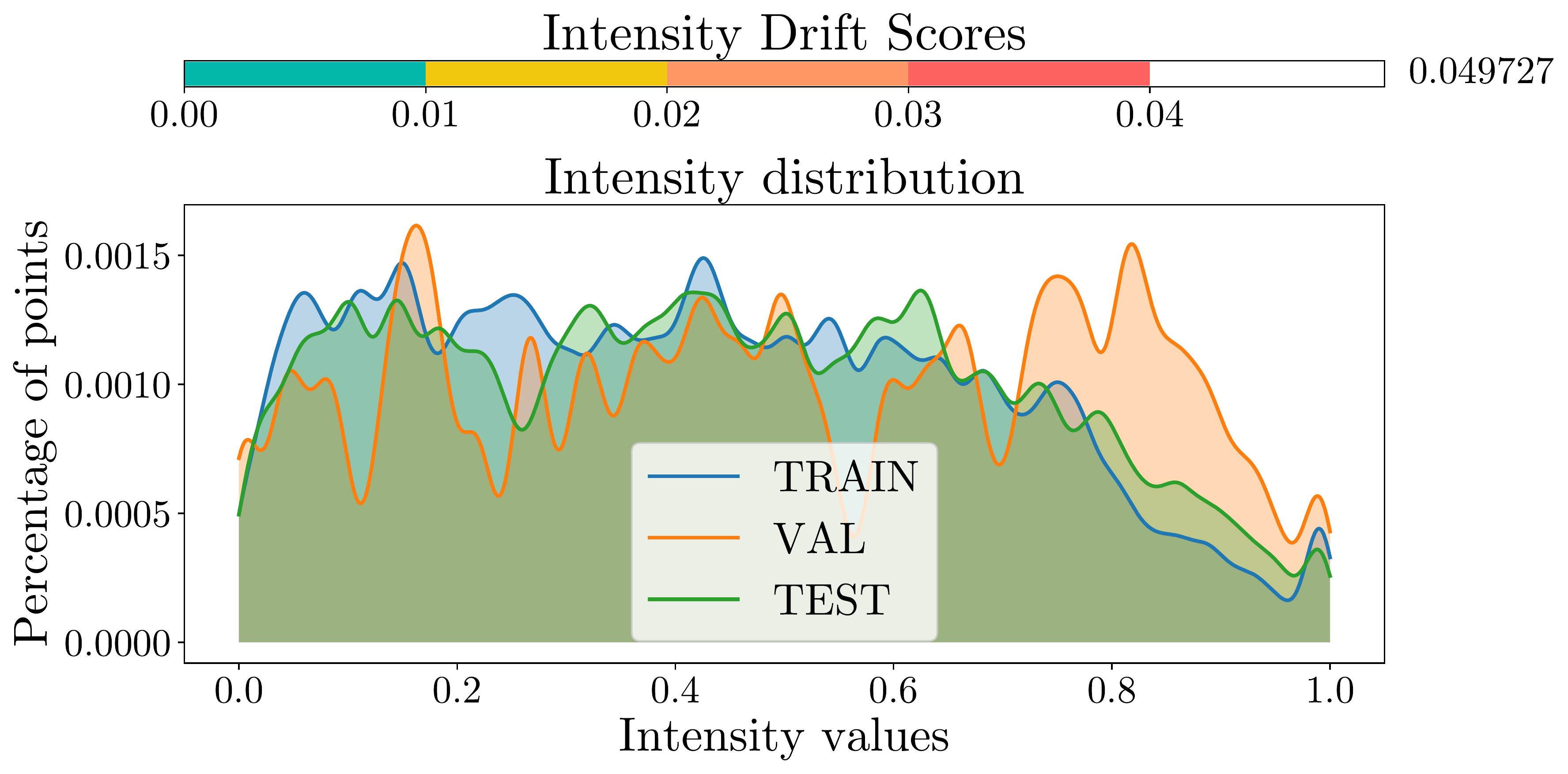} }}
    \caption{Intensity distribution drift.}
    \label{fig:intensity_drift}
\end{figure}

The figure \ref{fig:intensity_drift} shows the intensity distribution of each subset and the IDS corresponding to the splits of both Navya3DSeg and SemanticKITTI. These two datasets were built with different LiDAR sensors, from two different manufacturers. It explains the different shapes of the intensity distributions between the two datasets. Compared to SemanticKITTI, the selected split of Navya3DSeg is less prone to intensity distribution shift.

\begin{figure}[!h]
    \centering
    \subfloat[]{\includegraphics[width=\linewidth]{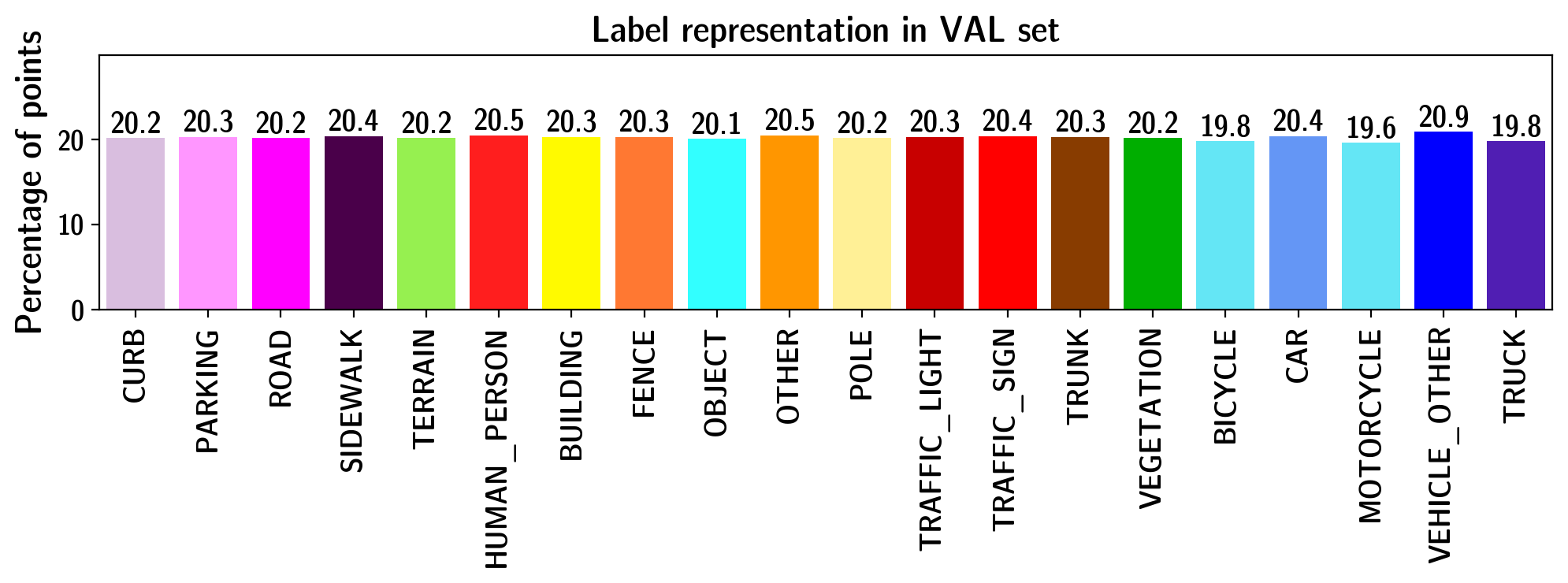}} \\
    \subfloat[]{\includegraphics[width=\linewidth]{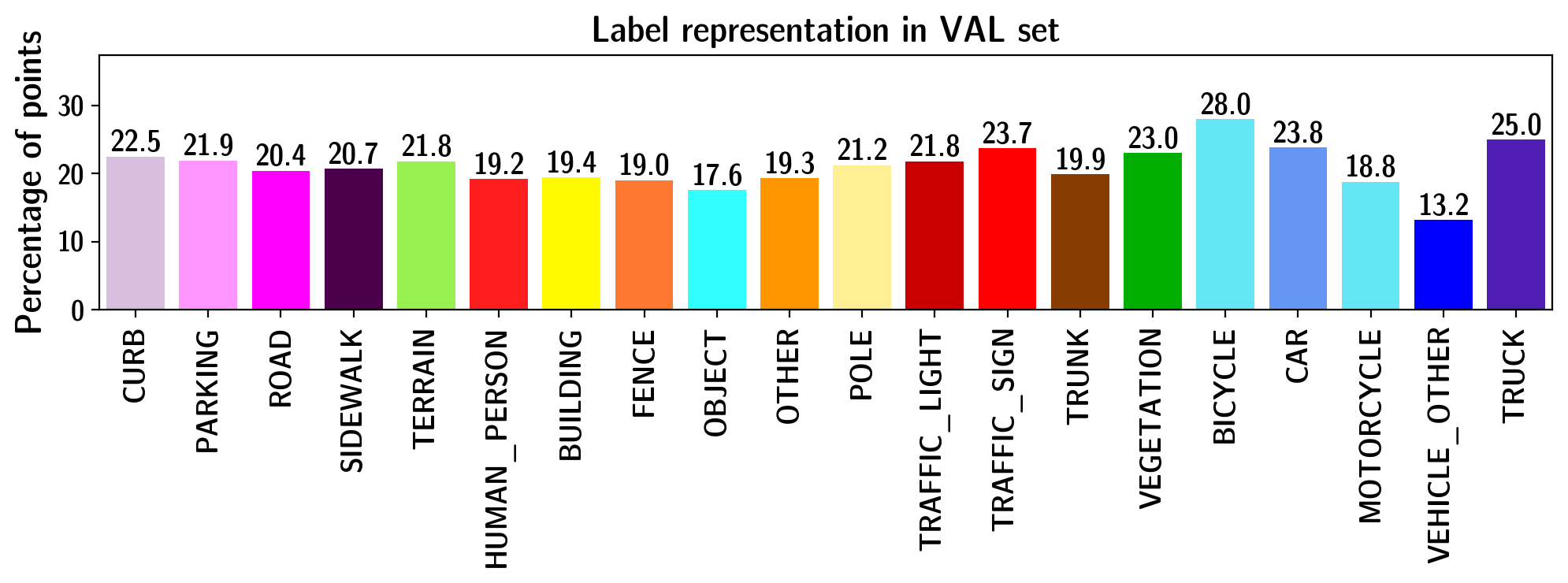}} \\
    \subfloat[]{\includegraphics[width=\linewidth]{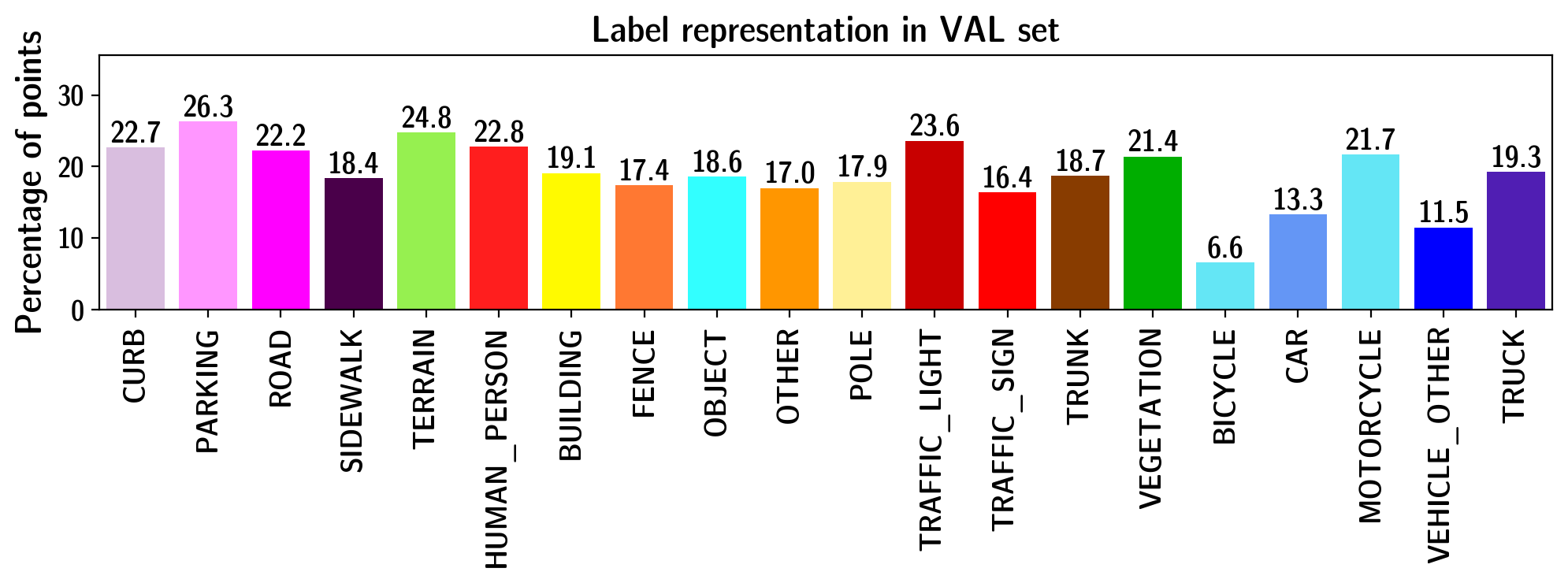}} \\

    \caption{Label representation of validation sets split by granularity of 1 (a), 500 (b) and 2000 (c) on Navya3DSeg}
    \captionsetup{justification=centering}
    \label{fig:split:granularity:n3ds:val_lp}
\end{figure}

\newpage

\subsection{Granularity experiments}
\begin{figure}[!h]
    \centering    
    \subfloat[]{\includegraphics[width=\linewidth]{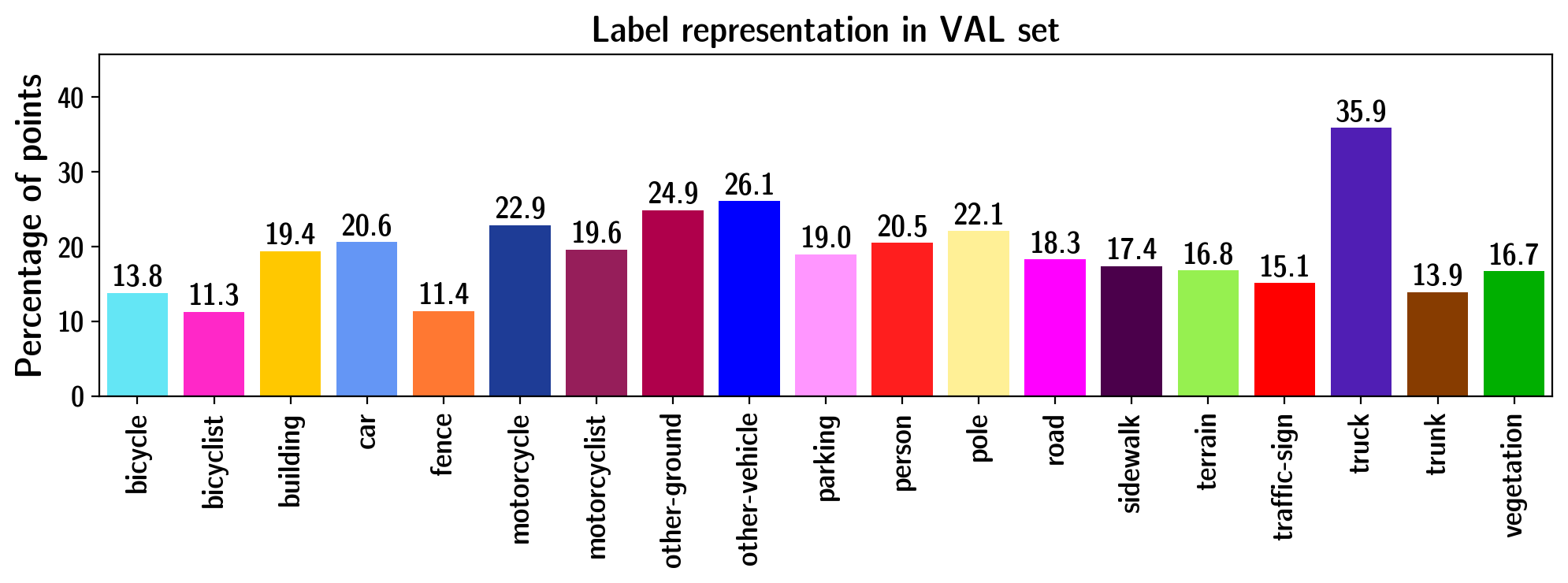}} \\
    \subfloat[]{\includegraphics[width=\linewidth]{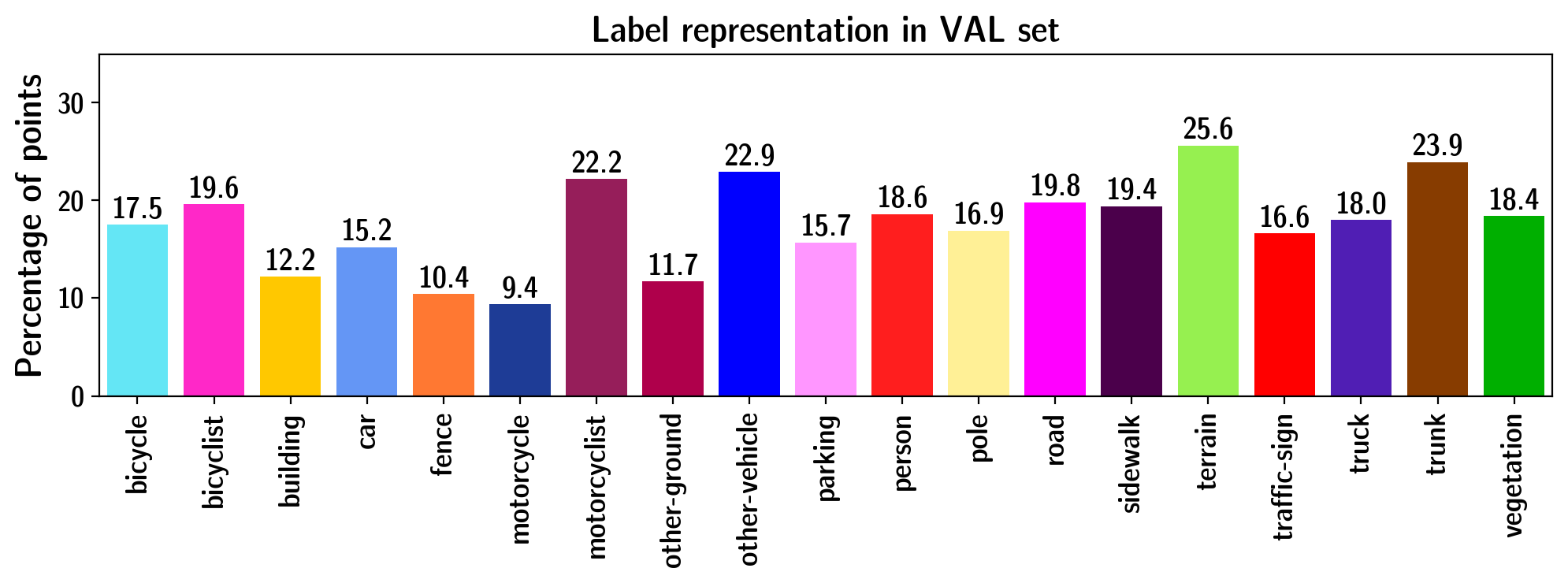}} \\
    \subfloat[]{\includegraphics[width=\linewidth]{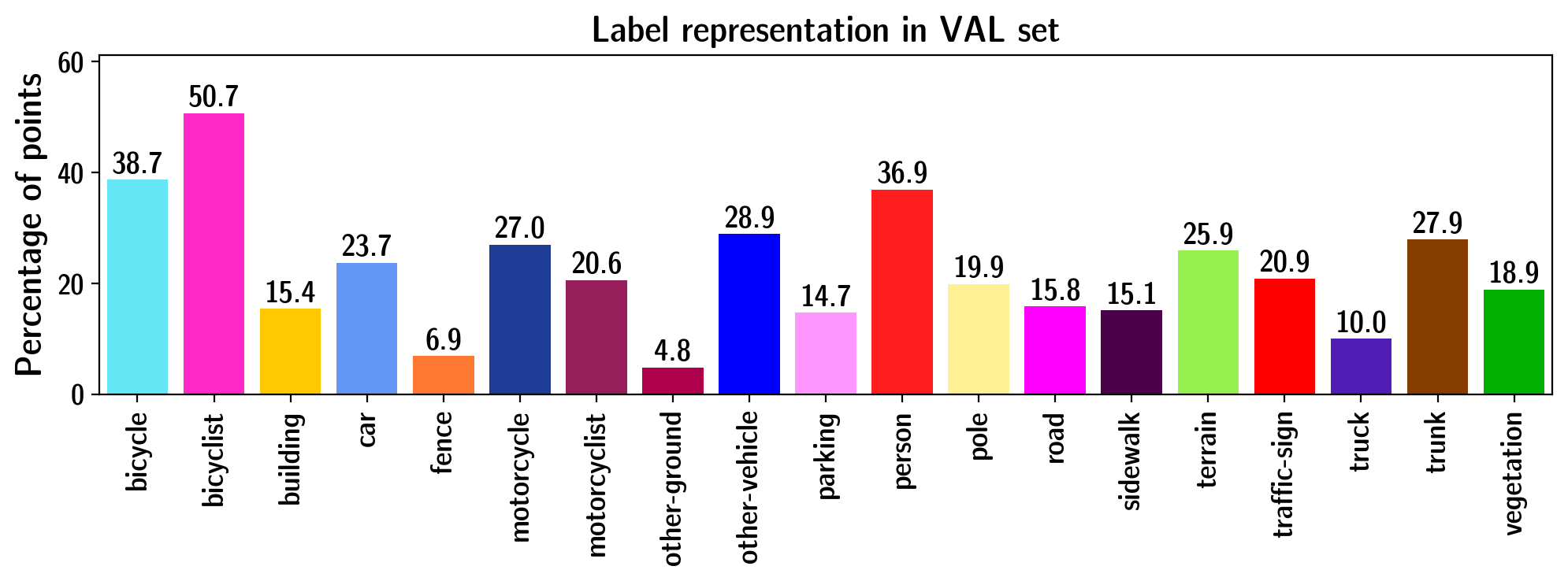}} \\
    \caption{Label representation of validation sets split by granularity of 500 (a) and 1000 (b) on SemanticKITTI and default validation set composed of sequence 08 (c). Dynamic labels are fused with static labels.}
    \captionsetup{justification=centering}
    \label{fig:split:granularity:sk:val_lp}
\end{figure}

Datasets containing sequential information such as point clouds sequences like in Navya3DSeg, or object tracking data in images could use the granularity parameter to define the level of aggregation to group individual samples. It allows advanced configuration to perform stratified sampling, with the Multi-label Stratified Shuffle Split (MSSS) technique, or our Multi-label Segment Stratified Shuffle Split (MSegSSS) technique. A reminder of our granularity benchmark is available at Tab. \ref{tab:appendix:split:granularity_exps}. The label distributions among subsets are presented in Fig. \ref{fig:split:granularity:n3ds:val_lp} for Navya3DSeg, and in Fig. \ref{fig:split:granularity:sk:val_lp} for SemanticKITTI.

\begin{table*}[ht]
\centering
\captionsetup{justification=centering}
\caption{Granularity evaluation on Navya3DSeg and SemanticKITTI. The sequence split (Sequence*) of SemanticKITTI was taken from the dataset configuration}
\scalebox{0.96}{
\begin{tabular}{c|cc|ccccccc}
\multirow{2}{*}{Dataset}       & \multicolumn{2}{c|}{Input variables} & \multicolumn{7}{c}{Output metrics}                                           \\ \cline{2-10} 
                               & Granularity     & Targeted ratios    & Obtained ratios    & LD $\downarrow$      & IDS $\downarrow$   & KL Divergence $\downarrow$    & Train mIoU & Val mIoU & Test mIoU \\ \hline
\multirow{5}{*}{Navya3DSeg}          & Sequence        & 87.5:12.5          & \textbf{79.8:20.2} & 5,10E-03 & 5,90E-03 & 6,45E-03 & 73.15        & 45.67    & 50.69     \\
 & 2000 & 87.5:12.5          & 87.3:12.7 & 4,60E-03 & 1,80E-02 & 7,01E-03 & 69.03 & 47.27 & 48.30 \\
 & $2000^\dagger$ & \textbf{79.8:20.2} & 79.5:20.5 & 4,30E-03 & 6,00E-03 & 4,27E-03 & 72.05 & 47.93 & 49.37 \\
 & 500  & \textbf{79.8:20.2} & 79.0:21.0 & 4,70E-03 & 3,60E-03 & 1,62E-03 & 72.77 & 54.59 & 50.49 \\
 & 1    & \textbf{79.8:20.2} & 79.7:20.3 & \textbf{2,00E-04} & \textbf{1,00E-04} & \textbf{7,00E-06} & 72.10 & 72.72 & \textbf{51.09} \\ \hline
\multirow{3}{*}{SemanticKITTI} & Sequence*         & -                  & \textbf{82.7:17.3} & 8,10E-03 & 6,53E-02 & 1,87E-02 & 80.84        & 53.66    & 54.81     \\
 & 1000 & \textbf{82.7:17.3} & 82.0:18.0 & 6,80E-03 & 3,50E-03 & 1,31E-02 & 81.62 & 61.43 & \textbf{56.00} \\
 & 500  & \textbf{82.7:17.3} & 82.5:17.5 & \textbf{3,40E-03} & \textbf{2,70E-03} & \textbf{4,35E-03} & 82.47 & 64.03 & 55.97
\end{tabular}}
\label{tab:appendix:split:granularity_exps}
\end{table*}

\newpage

\section{Semantic segmentation benchmark}

\subsection{Hyper-parameters \& Training setup}
The table \ref{tab:benchmark:train_params} shows the hyper-parameters and training set up for all experiments on Navya3DSeg, including dataset splitting, semantic segmentation benchmark, cross-dataset generalization, and dataset distillation with active learning. For the cyclic learning rate, we apply the implementation from PyTorch \cite{https://doi.org/10.48550/arxiv.1912.01703}. 
To perform an efficient benchmark, and lower the effect of label imbalance, we perform a label mapping, described in Tab. \ref{tab:label_mapping}, to gather some labels.

\begin{table}[!h] 
\centering
\captionsetup{justification=centering}
\caption{Models' hyper-parameters of training used in our different experiments. The \textit{triangular1.5} and \textit{triangular1.25} policies are similar to the standard \textit{triangular} policy, but instead it cuts our max learning rate bound in 1.5 and 1.25 respectively, after every cycle}
\scalebox{0.9}{
\begin{tabular}{c|cccc}
\textbf{Model}              & \multicolumn{1}{c|}{\textbf{weight decay}} & \multicolumn{1}{c|}{\textbf{Momentum}} & \multicolumn{1}{c|}{\textbf{Batch size}} & \textbf{Epochs} \\ \hline
SalsaNext                   & \multicolumn{1}{c|}{0,0001}                & \multicolumn{1}{c|}{0,9}               & \multicolumn{1}{c|}{36}                  & 150             \\ \cline{2-5} 
\&                          & \multicolumn{4}{c}{\textbf{Cyclic Learning rate (LR) scheduler\cite{https://doi.org/10.48550/arxiv.1912.01703} \cite{https://doi.org/10.48550/arxiv.1506.01186}}}                                                                                 \\ \cline{2-5} 
SqueezeSegV2                & \multicolumn{1}{c|}{Policy}                & \multicolumn{1}{c|}{Max LR}            & \multicolumn{1}{c|}{Base LR}             & Step size       \\ \cline{2-5} 
                            & \multicolumn{1}{c|}{Triangular 1,25}       & \multicolumn{1}{c|}{0,1}               & \multicolumn{1}{c|}{0,001}               & 8 epochs        \\ \hline
\multirow{5}{*}{Cylinder3D} & \multicolumn{1}{c|}{\textbf{weight decay}} & \multicolumn{1}{c|}{\textbf{Momentum}} & \multicolumn{1}{c|}{\textbf{Batch size}} & \textbf{Epochs} \\ \cline{2-5} 
                            & \multicolumn{1}{c|}{0,0001}                      & \multicolumn{1}{c|}{0.9}                  & \multicolumn{1}{c|}{10}                    & 150                 \\ \cline{2-5} 
                          & \multicolumn{4}{c}{\textbf{Cyclic Learning rate (LR) scheduler\cite{https://doi.org/10.48550/arxiv.1912.01703} \cite{https://doi.org/10.48550/arxiv.1506.01186}}}                                                                                 \\ \cline{2-5} 
                & \multicolumn{1}{c|}{Policy}                & \multicolumn{1}{c|}{Max LR}            & \multicolumn{1}{c|}{Base LR}             & Step size       \\ \cline{2-5} 
                            & \multicolumn{1}{c|}{Triangular 1,25}       & \multicolumn{1}{c|}{0,001}               & \multicolumn{1}{c|}{0,00001}               & 8 epochs        \\ \hline                
\end{tabular}
}
\label{tab:benchmark:train_params}
\end{table}

\subsection{Benchmark results}
\begin{figure}[]
    \centering
    \subfloat[]{\includegraphics[width=\linewidth]{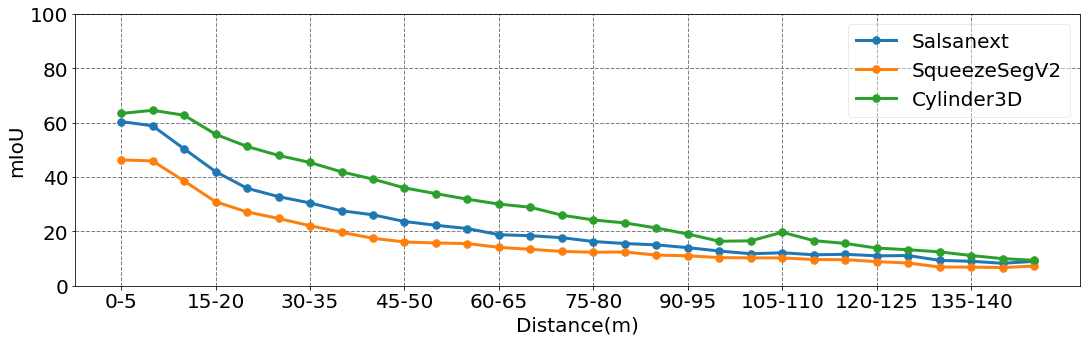}} \\
    \subfloat[]{\includegraphics[width=\linewidth]{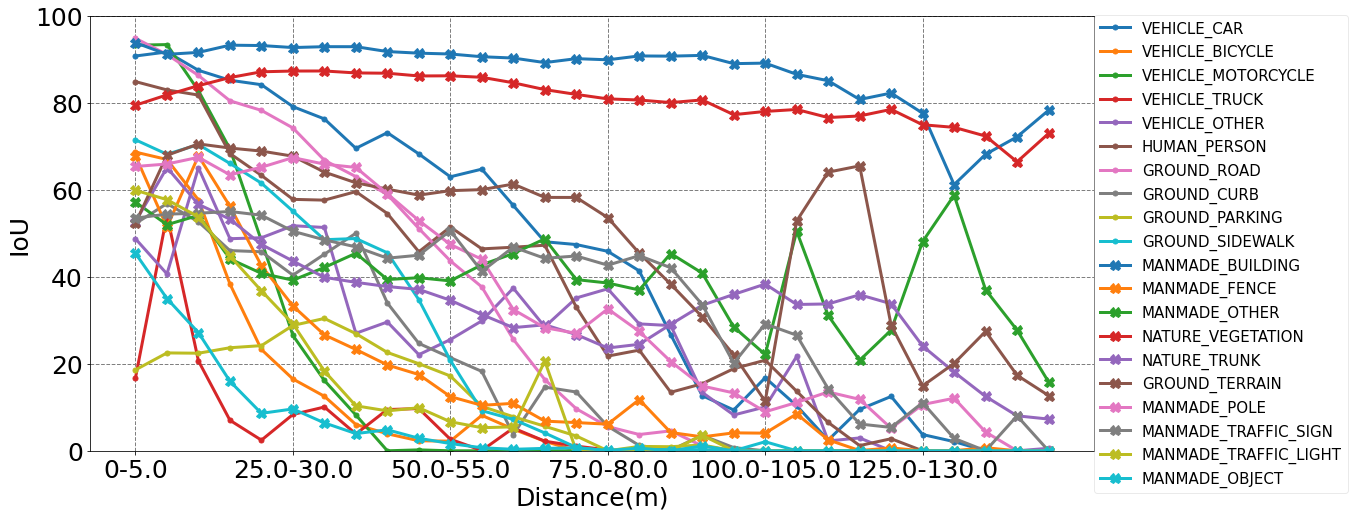}} \\
    \caption{(a): mIoU vs Distance plots for our trained models from 3 different architectures. (b): Label IoU plots for the Cylinder3D model.}
    \captionsetup{justification=centering}
    \label{fig:iou_vs_Dist}
\end{figure}

The results from our semantic segmentation benchmark are presented in Tab. \ref{tab:semseg_bench}. Overall, Cylinder3D performs better than the range image based architectures SqueezeSegV2 and SalsaNext. We plot the performance of these three baseline models over different ranges of operation in Fig. \ref{fig:iou_vs_Dist}.

\subsection{Cross-dataset generalization between Navya3DSeg and SemanticKITTI}

\begin{figure*}[]
    \centering
    \subfloat[\centering Navya3DSeg ROAD]{{\includegraphics[width=0.24\textwidth]{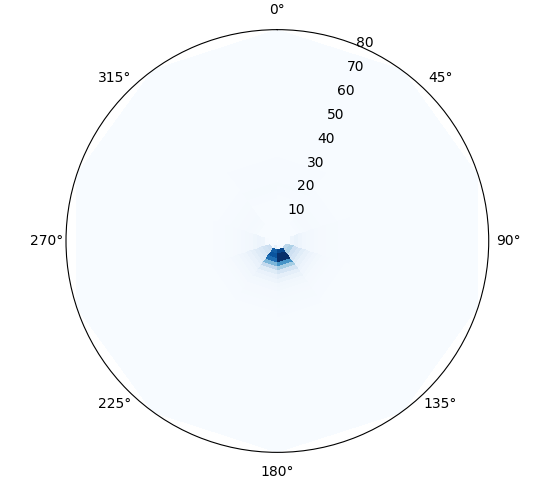}}}\hspace{1mm}
    \subfloat[\centering Navya3DSeg BUILDING]{{\includegraphics[width=0.24\textwidth]{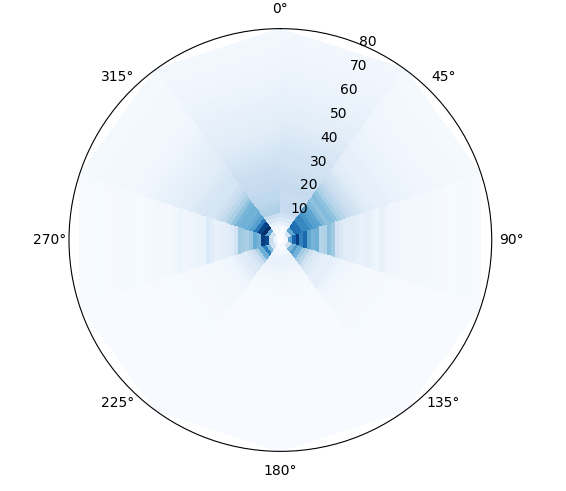}}}\hspace{1mm}
    \subfloat[\centering Navya3DSeg TERRAIN]{{\includegraphics[width=0.24\textwidth]{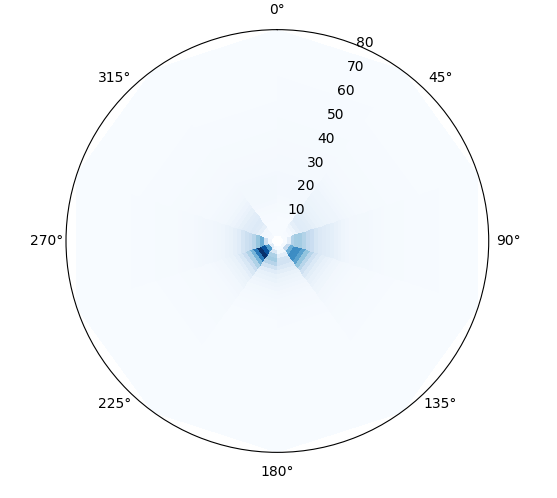}}}

    \subfloat[\centering SemanticKITTI ROAD]{{\includegraphics[width=0.24\textwidth]{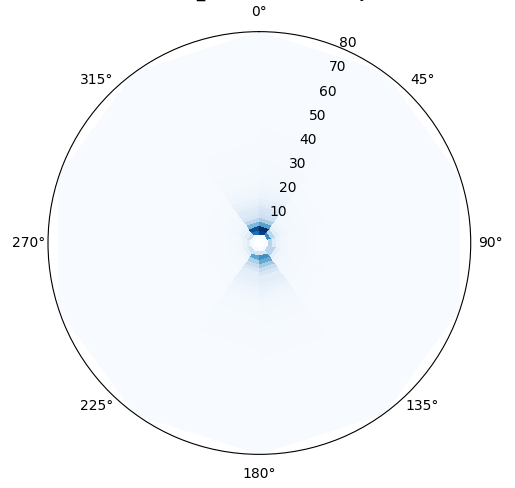}}}
    \hspace{1mm}
    \subfloat[\centering SemanticKITTI BUILDING]{{\includegraphics[width=0.24\textwidth]{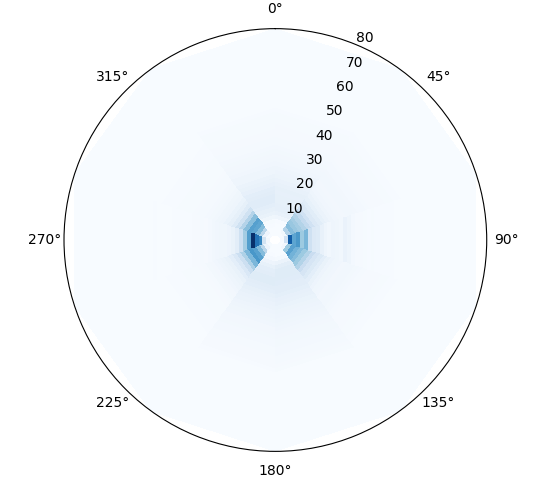}}}
    \hspace{1mm}
    \subfloat[\centering SemanticKITTI TERRAIN]{{\includegraphics[width=0.24\textwidth]{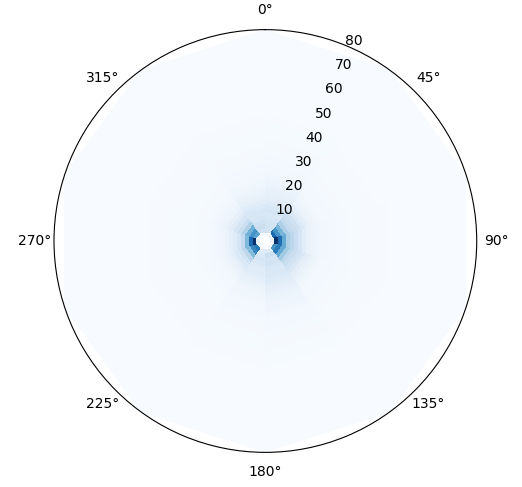}}}

    \subfloat[\centering Navya3DSeg TERRAIN]{{\includegraphics[width=0.24\textwidth]{images/polar_density_plot/n3ds_GROUND_TERRAIN.png}}}\hspace{1mm}
    \subfloat[\centering Navya3DSeg VEGETATION]{{\includegraphics[width=0.24\textwidth]{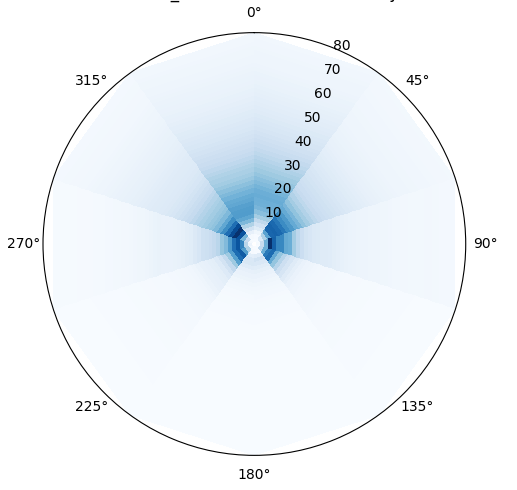}}}\hspace{1mm}
    \subfloat[\centering Navya3DSeg TRUNK]{{\includegraphics[width=0.24\textwidth]{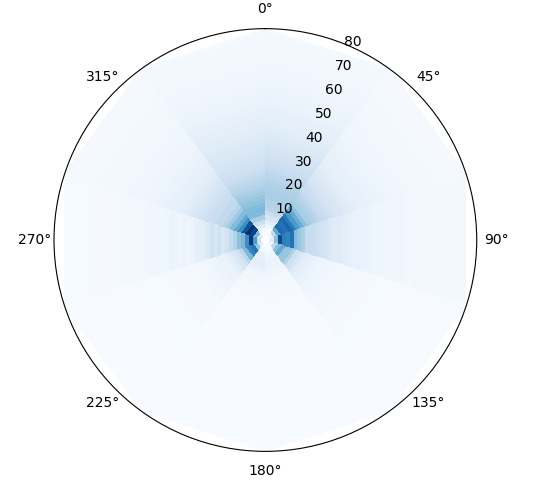}}}\hspace{1mm}
        
    \subfloat[\centering SemanticKITTI TERRAIN]{{\includegraphics[width=0.24\textwidth]{images/polar_density_plot/sk_GROUND_TERRAIN.png}}}
    \hspace{1mm}
    \subfloat[\centering SemanticKITTI VEGETATION]{{\includegraphics[width=0.24\textwidth]{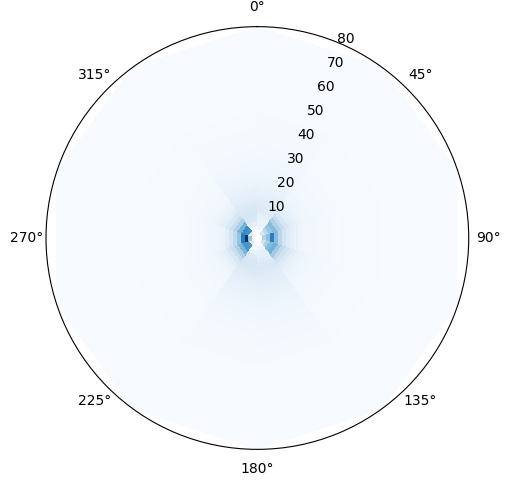}}}\hspace{1mm}
    \subfloat[\centering SemanticKITTI TRUNK]{{\includegraphics[width=0.24\textwidth]{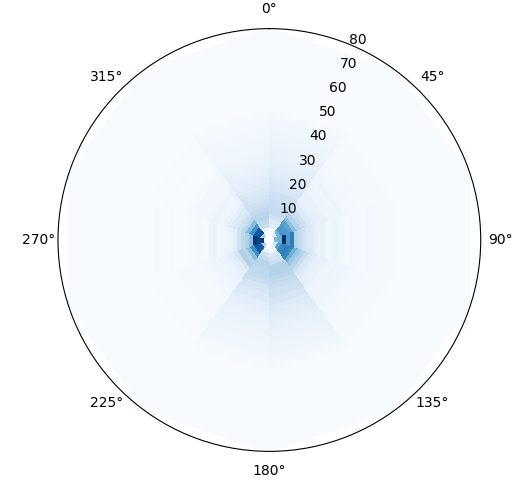}}}

    \subfloat[\centering Navya3DSeg CAR]{{\includegraphics[width=0.24\textwidth]{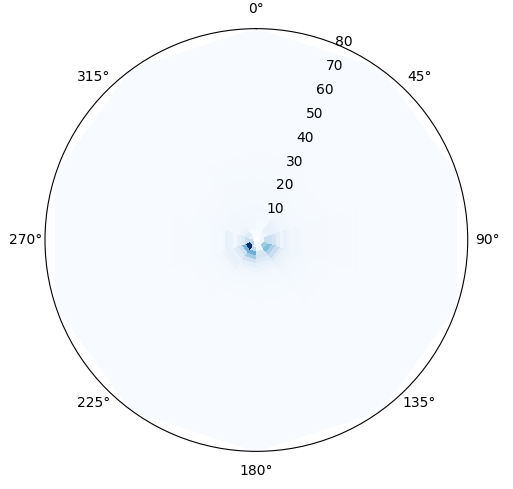}}}
    \hspace{1mm}
    \subfloat[\centering SemanticKITTI CAR]{{\includegraphics[width=0.24
    \textwidth]{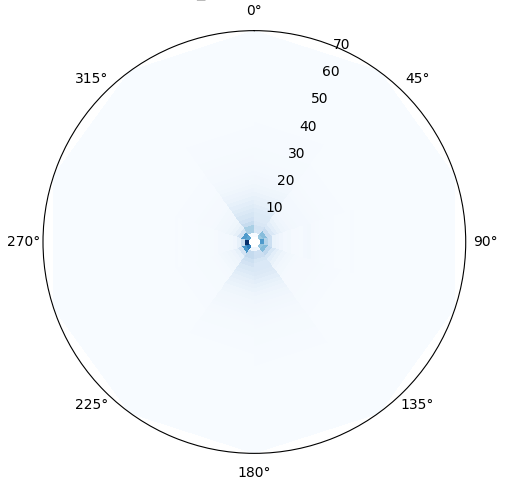}}}

    \caption{Polar density plot of highlighted labels for cross dataset generalisation experiment between Navya3DSeg and SemanticKITTI. The maximum radius has been set following SemanticKITTI annotation distance. Those plots emphasize the impact of Navya3DSeg sensor orientation compare to a flat configuration in SemanticKITTI.}
    \captionsetup{justification=centering}
    \label{fig:cdg:polar}
\end{figure*}

\begin{figure*}[]
    \centering
    \subfloat[\centering Navya3DSeg ROAD]{{\includegraphics[width=0.45\textwidth]{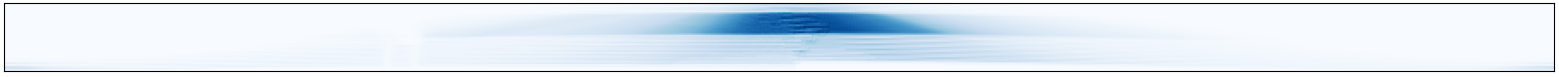}}}\hspace{1mm}
    \subfloat[\centering SemanticKITTI ROAD]{{\includegraphics[width=0.45\textwidth]{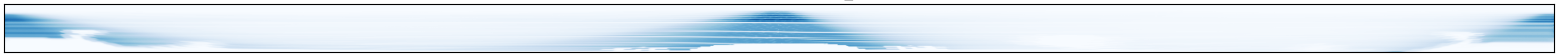}}}\hspace{1mm}

    \subfloat[\centering Navya3DSeg BUILDING]{{\includegraphics[width=0.45\textwidth]{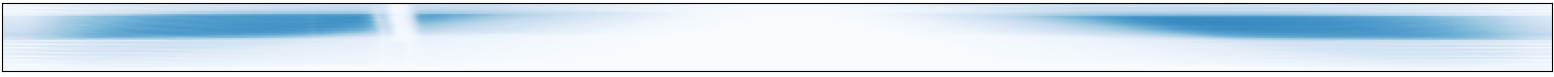}}}\hspace{1mm}
    \subfloat[\centering SemanticKITTI BUILDING]{{\includegraphics[width=0.45\textwidth]{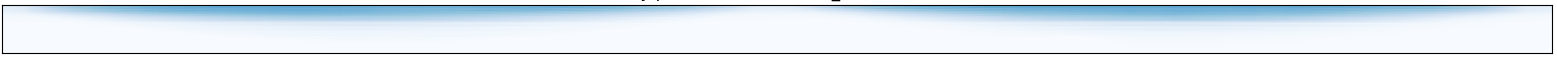}}}\hspace{1mm}

    \subfloat[\centering Navya3DSeg TERRAIN]{{\includegraphics[width=0.45\textwidth]{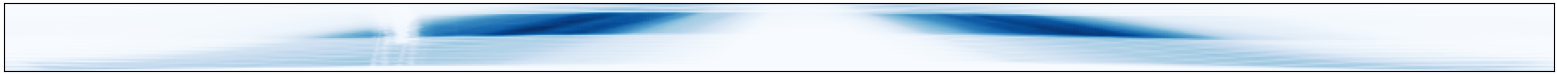}}}\hspace{1mm}
    \subfloat[\centering SemanticKITTI TERRAIN]{{\includegraphics[width=0.45\textwidth]{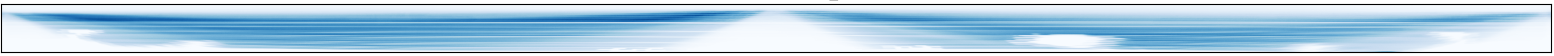}}}\hspace{1mm}

    \subfloat[\centering Navya3DSeg VEGETATION]{{\includegraphics[width=0.45\textwidth]{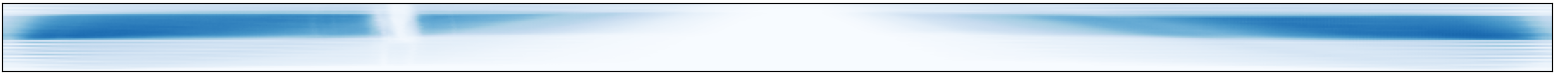}}}\hspace{1mm}
    \subfloat[\centering SemanticKITTI VEGETATION]{{\includegraphics[width=0.45\textwidth]{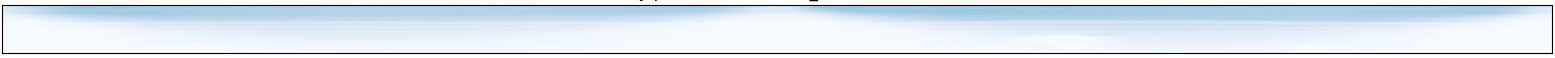}}}\hspace{1mm}

    \subfloat[\centering Navya3DSeg TRUNK]{{\includegraphics[width=0.45\textwidth]{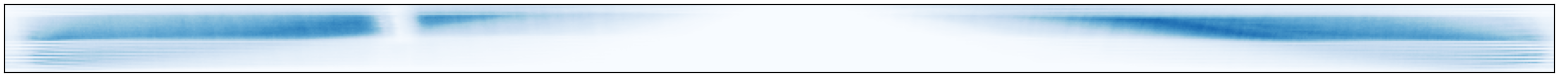}}}\hspace{1mm}
    \subfloat[\centering SemanticKITTI TRUNK]{{\includegraphics[width=0.45\textwidth]{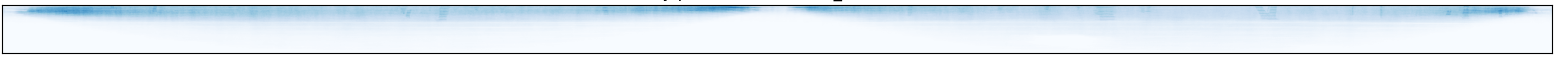}}}\hspace{1mm}

    \subfloat[\centering Navya3DSeg CAR]{{\includegraphics[width=0.45\textwidth]{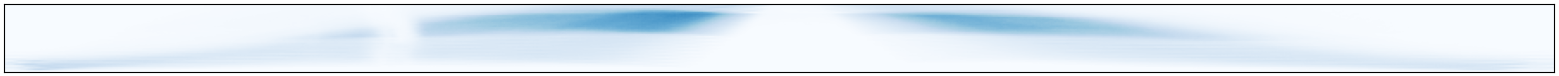}}}\hspace{1mm}
    \subfloat[\centering SemanticKITTI CAR]{{\includegraphics[width=0.45\textwidth]{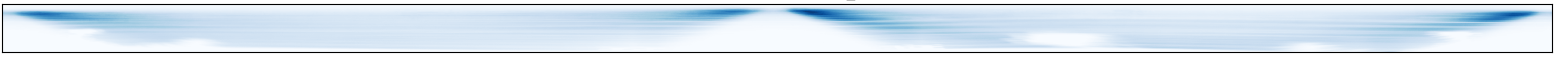}}}\hspace{1mm}

    \caption{Density range-images of highlighted labels for cross dataset generalisation experiment between Navya3DSeg and SemanticKITTI. No point limit has been set for spherical projection. Those plots shows the impact of spherical projection based on inclined sensor data.}
    \captionsetup{justification=centering}
    \label{fig:cdg:ri}
\end{figure*}

We extracted a common label space (Tab. \ref{tab:cross_dataset_kernel}) containing only labels available in both datasets' validation set and showing consistent annotation policies. For this experiment we performed the source-target label map dynamically. Because of SemanticKITTI's size and point cloud density, we expected to have better results when training on SemanticKITTI and evaluating on Navya3DSeg. However, Tab. \ref{tab:SSR:crossdatagen} shows that models trained on Navya3DSeg perform better on \textit{things} labels and on most of the ground labels, except GROUND\_ROAD, and MANMANDE\_BUILDING which are in the top-3 most represented class in SemanticKITTI, explaining the \textit{mIoU\_Stuff} score drop during evaluation. This can be explained by: 1. the Navya3DSeg sensor orientation which enhance the ground focus, and 2. by the spherical representation with this orientation, which distorts the projected range-image. Thus, when performing evaluation from Navya3DSeg, labeled points from the front and the rear of the mapping vehicle are highly impacted, contrary to the labeled points on the sides of the mapping vehicle. These differences in label distribution are highlighted in Fig. \ref{fig:cdg:polar} and Fig. \ref{fig:cdg:ri}.

\begin{table*}[]
\centering
\captionsetup{justification=centering}
\caption{Cross dataset generalization results between SemanticKITTI and Navya3DSeg on validation set with SalsaNext architecture}
\begin{tabular}{c|c|c|c|ccccccc}
Train dataset & Validation dataset            & \multicolumn{1}{c|}{mIoU Things} & \multicolumn{1}{c|}{mIoU Stuff} & \multicolumn{1}{c}{Road} & Building       & \multicolumn{1}{c}{Terrain} & \multicolumn{1}{c}{Vegetation} & \multicolumn{1}{c}{Trunk} & \multicolumn{1}{c}{Car} & \multicolumn{1}{c}{Traffic Sign} \\ \hline
SemanticKITTI & Navya3DSeg & 12.94                            & \textbf{22.82}                  & \textbf{46.72}            & \textbf{55.85} & 10.66                        & 40.39                           & 4.68                       & 21.42                    & 11.81                             \\ 
Navya3DSeg & SemanticKITTI & \textbf{17.55}                   & 16.36                           & 13.89                     & 17.71          & \textbf{14.17}               & \textbf{52.45}                  & \textbf{14.67}             & \textbf{29.40}           & \textbf{22.14}                    \\
\end{tabular}
\label{tab:SSR:crossdatagen}
\end{table*}

\begin{table}[]
\captionsetup{justification=centering}
\caption{Semantic segmentation benchmark on Navya3DSeg test and validations sets}
\resizebox{\columnwidth}{!}{
\begin{tabular}{c|c|cc|cc|c|c}
\multirow{2}{*}{Subset} &
  \multirow{2}{*}{Model} &
  \multicolumn{2}{c|}{Things labels} &
  \multicolumn{2}{c|}{Stuff labels} &
  \multirow{2}{*}{mIoU} &
  \multirow{2}{*}{fwIoU} \\ \cline{3-6}
                            &              & mIoU           & fwIoU & mIoU           & fwIoU     &                &           \\ \hline
\multirow{3}{*}{Validation} & SalsaNext    & 43,73          & 1,41  & 46,50          & 69,48     & 45,67          & 70,88     \\
                            & SqueezeSegV2 & 27,87          & 1,22  & 35,10          & 59,67     & 32,93          & 60,89     \\
                            & Cylinder3D   &       \textbf{58.69}         &   \textbf{1.72}    &       \textbf{54.23}         &    \textbf{75.85}       &        \textbf{55.57}        &      \textbf{77.57}     \\ \hline
\multirow{3}{*}{Test}       & SalsaNext    & 50,18          & 2,25  & 50,90          & 71,86     & 50,69          & 74,11     \\
                            & SqueezeSegV2 & 37,71          & 1,93  & 40,12          & 65,59     & 39,39          & 67,52     \\
                            & Cylinder3D   & \textbf{68.40} &   \textbf{2.66}    & \textbf{60.46} & \textbf{78.50} & \textbf{62.84} & \textbf{81.16}
\end{tabular}
}
\label{tab:semseg_bench}
\end{table}

\section{Dataset distillation with Active Learning}

\subsection{ego-pose distance based sampling}
Ego-pose distance based sampling (detailed in Algo. \ref{alg:ds}) can be adapted either offline or online in an active learning setup. 
In terms of practical implementation, we use the \textit{BallTree} from \textit{sklearn.neighbors} to naively find neighbors within a radius, and euclidean distance as the distance between two point clouds. 

\begin{algorithm}[!h]
\SetKwInOut{Input}{Input}
\SetKwInOut{Output}{Output}
\SetKwComment{Comment}{/* }{ */}
\caption{Ego-pose distance based sampling proposed method}
\label{alg:ds}
\Input{labeled pool $L$, unlabeled pool $U$, budget $b$ }

\Comment{\scriptsize Compute the maximum distance between labeled and unlabeled samples' ego-poses}

\eIf{$L = \emptyset$}
{$d=$MaxDistance(U, U)}
{$d=$MaxDistance(L, U)}

\Comment{\scriptsize Find unlabeled samples the don't belong to the neighbors of L within a radius d}
$A = \{x | x \in U,  \forall y \in L,  \text{dist}(x, y) > d\}$

$S=\emptyset$\\
\For{$n \leftarrow 1$ \KwTo $b$}{
    \While {$A = \emptyset$} {
    d = reduce(d)    \Comment{\scriptsize reduce radius d}
    \Comment{\scriptsize Find unlabeled samples that don't belong to the neighbors of L and S within a radius d}
    $A = \{x | x \in U\setminus S,  \forall y \in L \sqcup S,  \text{dist}(x, y) > d\}$
    }

    \uIf{$S \ne \emptyset$}{
  
    \Comment{\scriptsize  select a sample in A having the maximum distance to its closest point in S.}
    $s \leftarrow \{x|\text{dist}(x, y)= \max_{x'}\text{dist}(x', y),~  \text{dist}(x', y)=\min_{y'}dist(x', y'),  \forall x'\in  A, \forall y' \in S \}$
    }
    \uElseIf{$L \ne \emptyset$}{
    \Comment{\scriptsize  select a sample in A having the maximum distance to its closest point in L.}
    $s \leftarrow \{x|\text{dist}(x, y)= \max_{x'}\text{dist}(x', y),~  \text{dist}(x', y)=\min_{y'}dist(x', y'),  \forall x'\in  A, \forall y' \in L \}$
    }
    \Else{
    $s \leftarrow $ randomly select a sample in A.
    }
    
    $S \leftarrow S \cup \{s\}$
    }
\Output{selected samples $S=\{s_1, ..., s_b\}$}

\end{algorithm}

\subsection{Active learning benchmark}

\textbf{Metrics}
 Given a specific value of \textit{mIoU}, the Labeling Efficiency (LE) based on \textit{mIoU} is the ratio between the number of labeled range images, acquired by the baseline sampling and the other sampling techniques.
    \begin{equation}
    \text{LE} = \frac{n_{\text{labeled}\_\text{others}}(\text{mIoU}=a)}{n_{\text{labeled}\_\text{baseline}}(\text{mIoU}=a)}    
    \end{equation}
    The baseline method in our experiment is the random heuristic.

\begin{figure*}[]
    \centering
    \includegraphics[width=\textwidth]{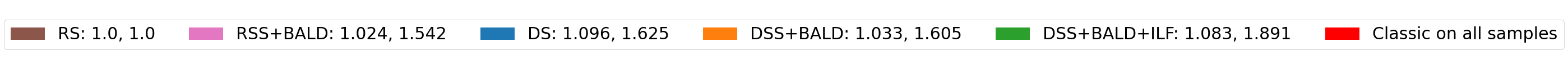}
    \subfloat[]{\includegraphics[height=4.5cm,width=0.48\textwidth]{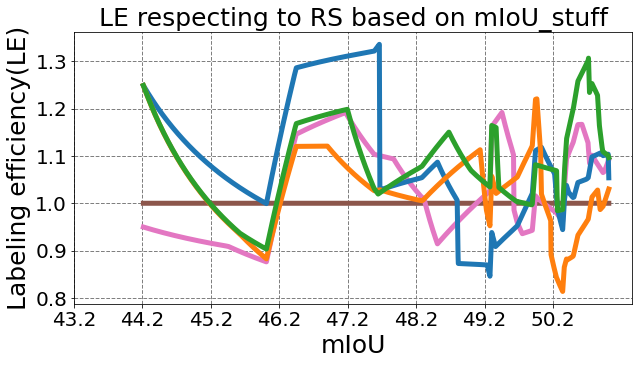}} \hfill
    \subfloat[]{\includegraphics[height=4.5cm,width=0.48\textwidth]{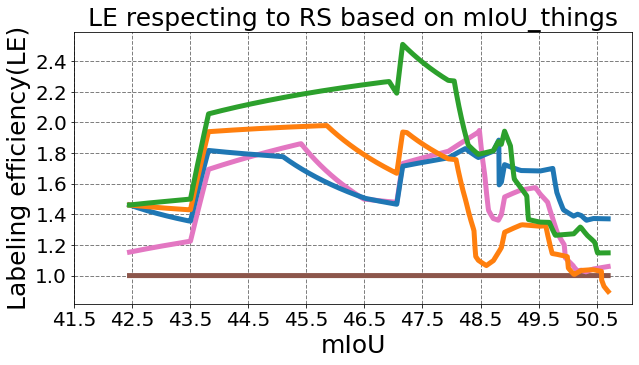}}
    \caption{Navya3DSeg dataset distillation results. Values on the legend are the averages of LE ratios based on \textit{stuff mIoU} (b) and \textit{things mIoU} (c). 
    }
    \captionsetup{justification=centering}
    \label{fig:al:le_miou_stuff_things}
\end{figure*}

\begin{figure*}[]
    \centering
    \subfloat[\centering sequence 08 scan 000305]{{\includegraphics[width=0.45\textwidth]{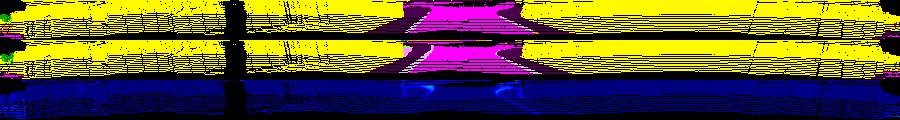}}}\hspace{1mm}
    \subfloat[\centering sequence 14 scan 001508]{{\includegraphics[width=0.45\textwidth]{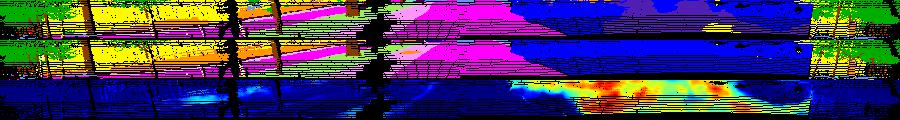}}}

    \subfloat[\centering sequence 08 scan 000299]{{\includegraphics[width=0.45\textwidth]{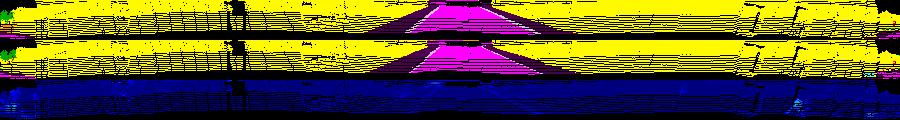}}}\hspace{1mm}
    \subfloat[\centering sequence 06 scan 003453]{{\includegraphics[width=0.45\textwidth]{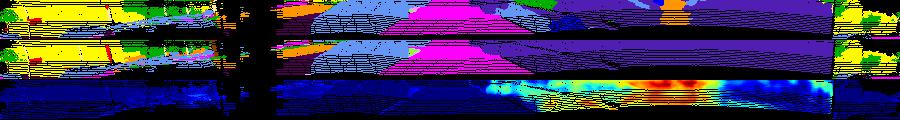}}}

    \subfloat[\centering sequence 02 scan 002503]{{\includegraphics[width=0.45\textwidth]{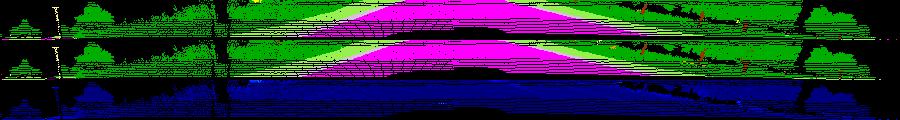}}}\hspace{1mm}
    \subfloat[\centering sequence 14 scan 002710]{{\includegraphics[width=0.45\textwidth]{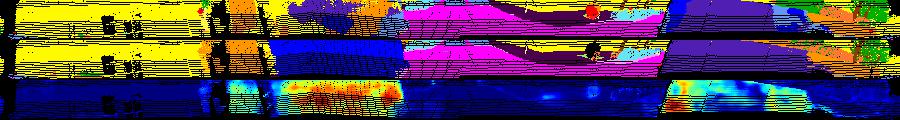}}}
    \caption{Active learning sample selection examples for DSS+BALD+ILF run. For each scan, from top to bottom, images are prediction, ground truth and BALD score. (a), (c) and (e): Top 3 easiest samples in subset. (b), (d) and (e): Top 3 hardest samples in subset.}
    \captionsetup{justification=centering}
    \label{fig:al:dss_bald_ilf_selection}
\end{figure*}

\begin{figure*}[]
    \centering
    \includegraphics[width=\textwidth]{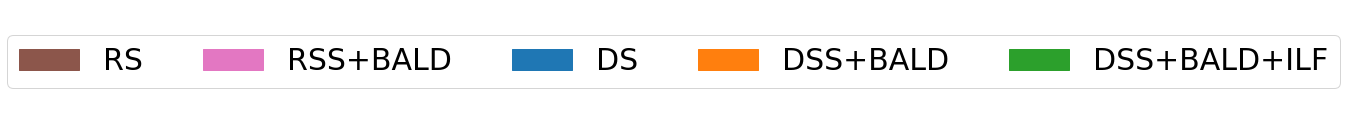}
    \includegraphics[width=0.48\textwidth]{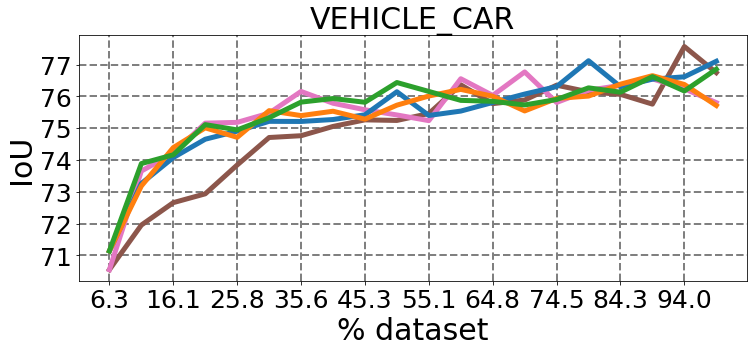}
    \includegraphics[width=0.48\textwidth]{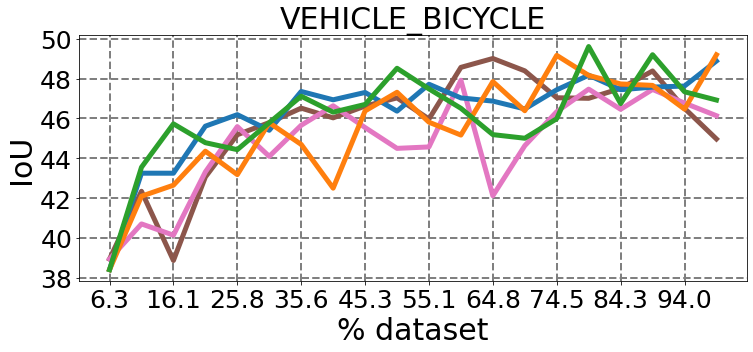}
    \includegraphics[width=0.48\textwidth]{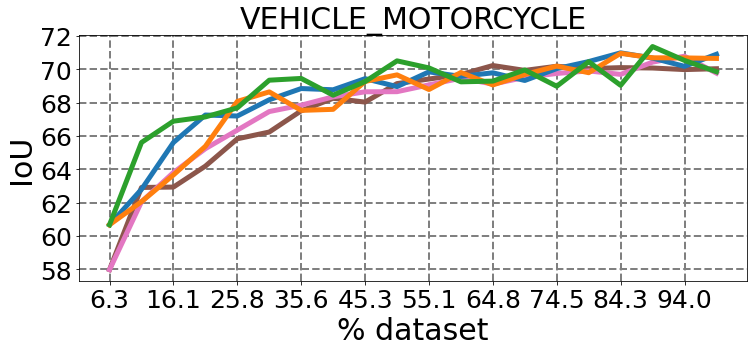}
    \includegraphics[width=0.48\textwidth]{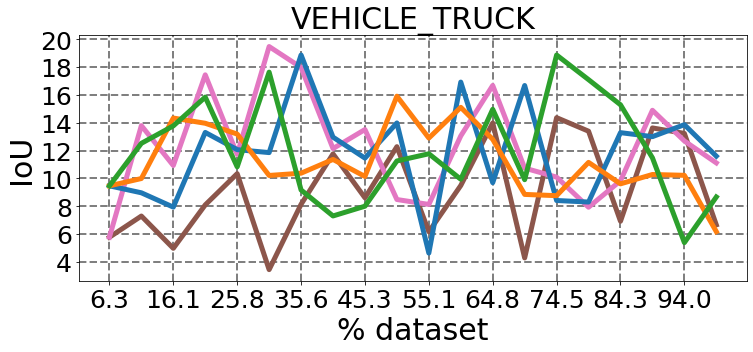}
    \includegraphics[width=0.48\textwidth]{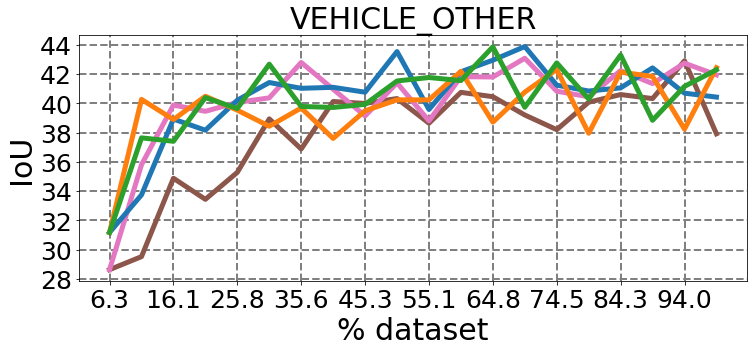}
    \includegraphics[width=0.48\textwidth]{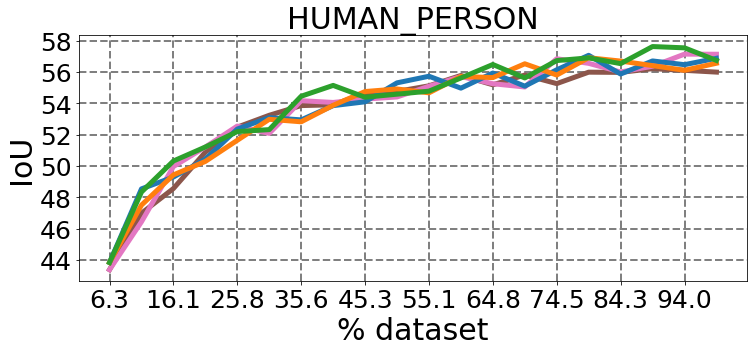}
    \includegraphics[width=0.48\textwidth]{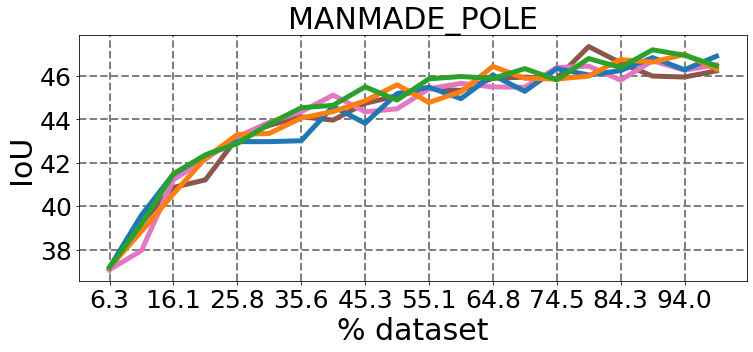}
    \includegraphics[width=0.48\textwidth]{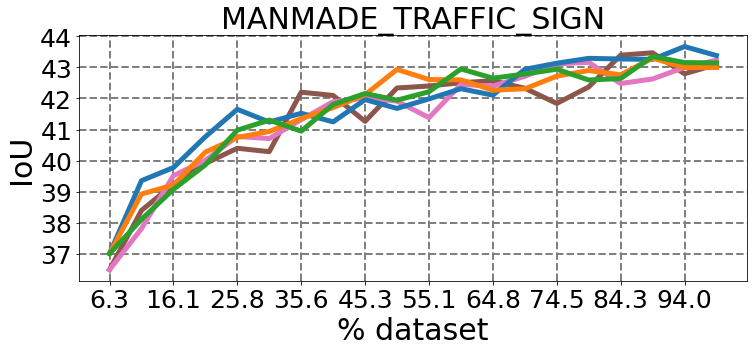}
    \includegraphics[width=0.48\textwidth]{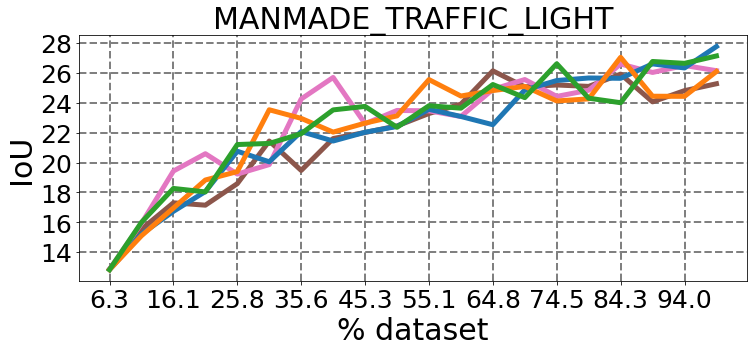}
    \includegraphics[width=0.48\textwidth]{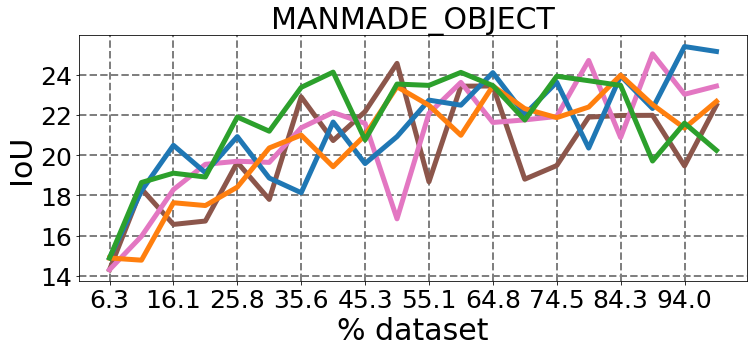}
    \caption{IoU vs \% dataset of \textit{things} labels.}
    \captionsetup{justification=centering}
    \label{fig:al:classiou:things}
\end{figure*}

\begin{figure*}[]
    \centering
    \vspace{-10mm}
    \includegraphics[width=\textwidth]{images/AL_per_class/legend.png} 
    \includegraphics[width=0.48\textwidth]{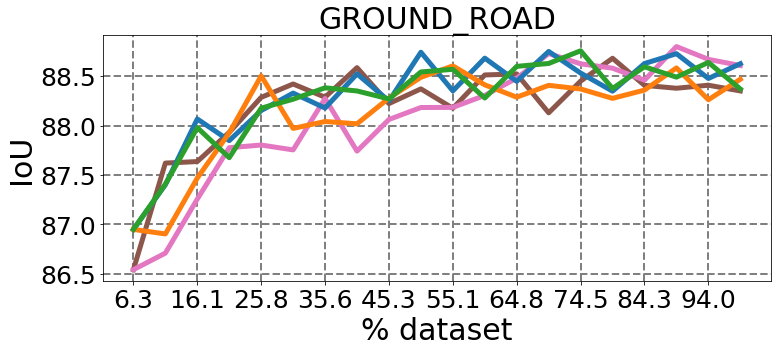}
    \includegraphics[width=0.48\textwidth]{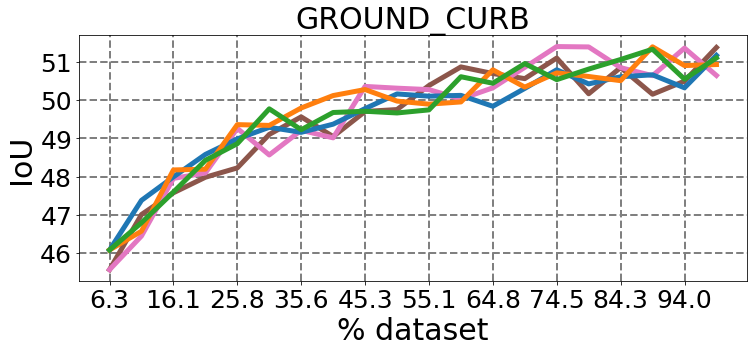}
    \includegraphics[width=0.48\textwidth]{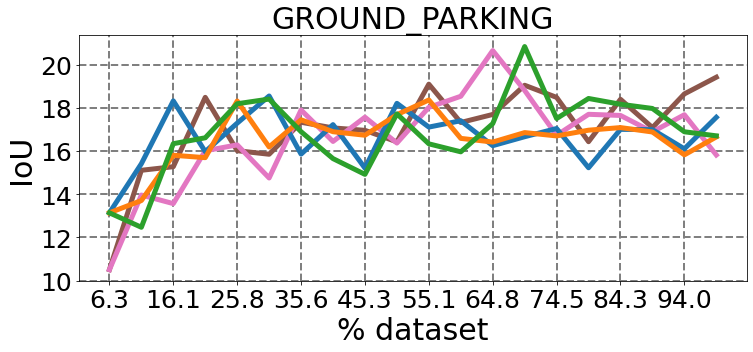}
    \includegraphics[width=0.48\textwidth]{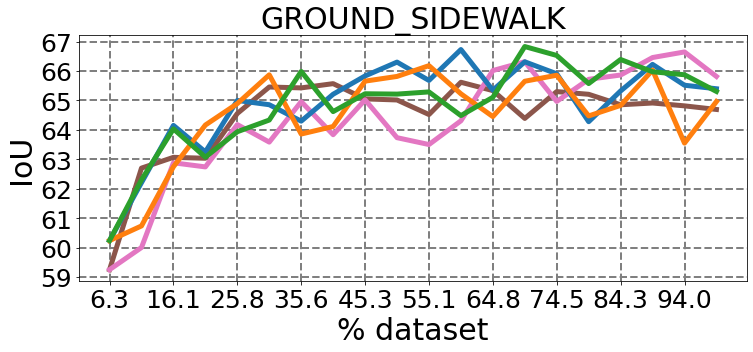}
    \includegraphics[width=0.48\textwidth]{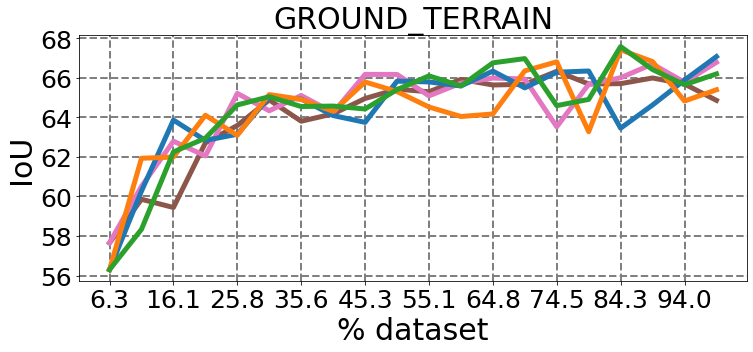}
    \includegraphics[width=0.48\textwidth]{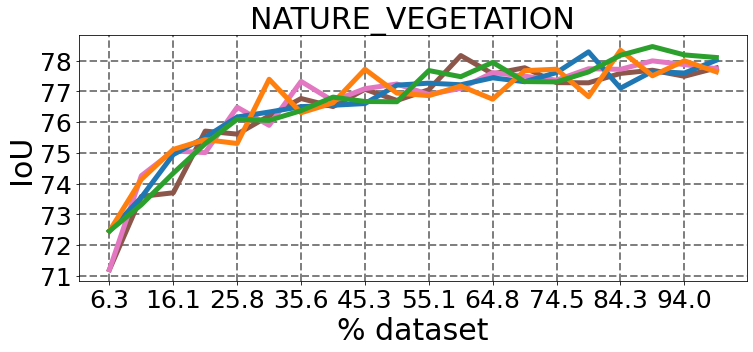}
    \includegraphics[width=0.48\textwidth]{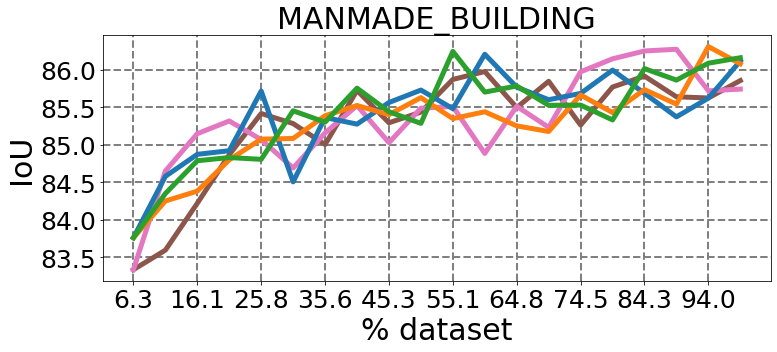}
    \includegraphics[width=0.48\textwidth]{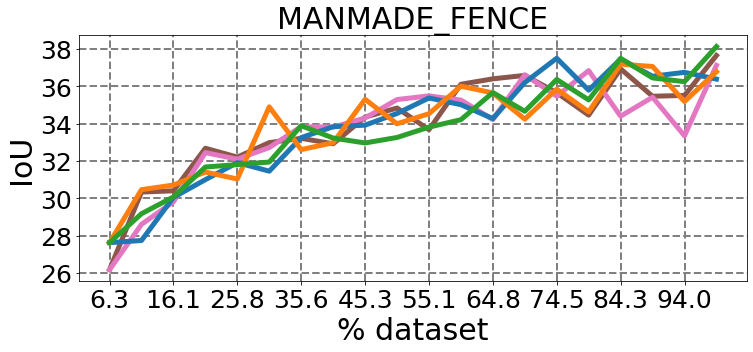}
    \includegraphics[width=0.48\textwidth]{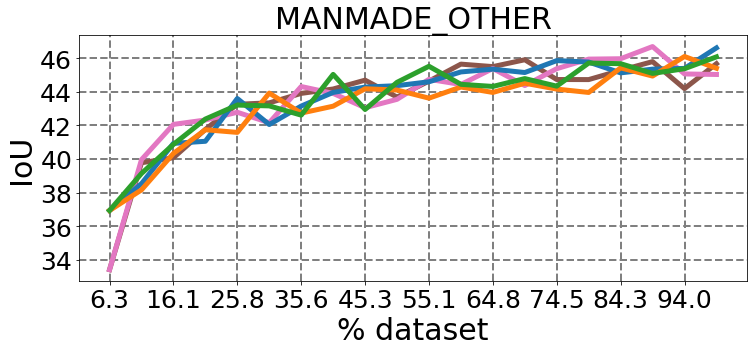}
    \includegraphics[width=0.48\textwidth]{images/AL_per_class/IoU_of_NATURE_VEGETATION_vs__dataset.png}
    \includegraphics[width=0.48\textwidth]{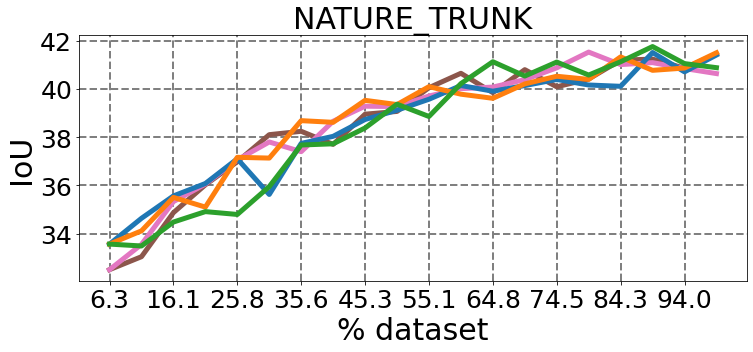}
    \caption{IoU vs \% dataset of \textit{stuff} labels.}
    \captionsetup{justification=centering}
    \label{fig:al:classiou:stuff}
\end{figure*}

\textbf{Extra results} According to Fig. \ref{fig:al:le_miou_stuff_things}, our ego-pose distance based sampling method gives the best performance on \textit{stuff} labels, whereas BALD and Inverse Label Frequency (IFL) focuses on \textit{things} labels. Among \textit{things} labels (Fig. \ref{fig:al:classiou:things}), VEHICLE\_CAR and VEHICLE\_OTHER have the most visible performance improvement between Random (RS) and the other methods. On the contrary, \textit{stuff} labels (Fig. \ref{fig:al:classiou:stuff}) are more stable for all methods, but their performances still increase as AL steps increase. Fig. \ref{fig:al:dss_bald_ilf_selection} shows the top 3 hardest and easiest selected samples of our best strategy (DSS+BALD+ILF).

\begin{table*}[]
\centering
\footnotesize
\caption{Navya3DSeg labels definition}
\label{tab:n3ds_Labels_definition}
\resizebox{\textwidth}{!}{
\begin{tblr}{Q[c,1cm]Q[l,3cm]Q[l,12cm]}
\SetCell[c=1]{c} Category & \SetCell[c=1]{c} Label & \SetCell[c=1]{c} Definition \\ \hline
\SetCell[r=2]{c} \rotatebox[origin=c]{90}{Other} & UNKNOWN & Undefined object or stuff \\ 
& OTHER & Outlier points, caused by reflections or from mapping vehicle track for example \\ \hline
\SetCell[r=5]{c} \rotatebox[origin=c]{90}{Vehicle} & CAR & 4-wheels cars, sedan, SUV, city car, vans with a continuous cabin \\
    & BICYCLE & 2-wheels bicycle without the rider \\
    & MOTORCYCLE & 2-wheels motorcycle without the rider \\
    & TRUCK & Trucks, vans with a body that is separated from the driver cabin, pickup trucks, with attached trailers. \\
    & OTHER & Any vehicle that does not fit into the other category's labels. Eg. bus, airplane, boat, another autonomous vehicle \\ \hline
\SetCell[r=1]{c} \rotatebox[origin=c]{90}{Human} & PERSON & Humans as pedestrians or riders of a vehicle \\ \hline
\SetCell[r=9]{c} \rotatebox[origin=c]{90}{Ground} & ROAD & Areas where motorized vehicles are supposed to move on \\
    & SPEED\_BUMP & Areas on the road materializing a speed bump, or only a slope intended to make a slow down \\
    & PARKING & Areas where vehicles, such as cars, bicycles, motorcycles and trucks are allowed to park. An area where a car is parked is not considered as parking by itself (it can be an illegal parking) \\
    & BICYCLE\_LANE & Areas dedicated to bicycles only. Must be delimited at least by a lane marking \\
    & CURB & Structures that delimits the slope between the road and the sidewalk \\
    & PEDESTRIAN\_CROSSING & Areas where the road marking is a zebra one, materializing a pedestrian crossing \\
    & ZEBRA\_ROAD\_MARKING & Areas with painted road markings, showing a clear “no go zone” for any vehicle \\
    & SIDEWALK & Areas dedicated to pedestrians only or areas dedicated to pedestrians and bicycles only. \\
    & TERRAIN & Areas with grass and all other types of horizontal spreading vegetation, including soil, where a human can normally walk on. Any high vegetation (height \textgreater $\sim$20cm) must be considered as vegetation \\
    & OTHER & Any kind of ground plane area that does not fit into the definitions of other ground labels \\ \hline
\SetCell[r=7]{c} \rotatebox[origin=c]{90}{Manmade} & BUILDING & Structures where a human can go inside. It can be a house, an office building, a skyscraper, a garage, a garden shed, … \\
    & FENCE & Structures on the yz-plane, which contains holes in it. Flat and solid yz-areas should be considered as walls, which can belong to a building or manmade\_other label. \\
    & OTHER & Any kind of structure that does not fit into any other Manmade labels. It includes fixed structures where a human cannot go inside, art sculptures, toll gate, … \\
    & POLE & Structures that can be considered as long, straight and vertical. It can be street lamp, electric pole, bollard or the structure meant to support traffic lights and signs. It also includes the structure at the top of the pole, like the street lamp itself. \\
    & TRAFFIC\_SIGN & Flat panels with traffic information written on it. Most of the time the surface is reflective. It can also be directions panels. \\
    & TRAFFIC\_LIGHT & Infrastructures to signal traffic with lights. Either 3-row traffic lights, or a single light. \\
    & OBJECT & Objects that can be moved by a single human. Eg. traffic cone, trash can, flower pot, a bag, etc \\ \hline
\SetCell[r=2]{c} \rotatebox[origin=c]{90}{Nature} & VEGETATION & Vegetation where a human cannot normally walk on, including bushes, foliage \\
    & TRUNK & Natural tree trunks, separated from the tree foliage \\ \hline
\SetCell[r=5]{c} \rotatebox[origin=c]{90}{Dynamic} & CAR & Dynamic tracks of vehicles belonging to the car label \\
    & BICYCLIST & Dynamic tracks of bicycle or motorcycle \\
    & PERSON & Dynamic tracks of human label \\
    & TRUCK & Dynamic tracks of  truck label \\
    & OTHER\_VEHICLE & Dynamic tracks of other\_vehicle label
\end{tblr}
}
\end{table*}


\begin{table*}[]
\centering
\captionsetup{justification=centering}
\caption{Benchmark label mapping from original Navya3DSeg to Navya3DSeg Benchmark with sub category information}
\label{tab:label_mapping}
\begin{tabular}{l|l|l}
\hline
\textbf{Navya3DSeg Label}    & \textbf{Benchmark Label} & \textbf{SubSet} \\ \hline
UNKNOWN                      & UNKNOWN                  & None            \\
OTHER                        & UNKNOWN                  & None            \\ \hline
VEHICLE\_CAR                 & VEHICLE\_CAR             & Thing           \\
VEHICLE\_BICYCLE             & VEHICLE\_BICYCLE         & Thing           \\
VEHICLE\_MOTORCYCLE          & VEHICLE\_MOTORCYCLE      & Thing           \\
VEHICLE\_TRUCK               & VEHICLE\_TRUCK           & Thing           \\
VEHICLE\_OTHER               & VEHICLE\_OTHER           & Thing           \\ \hline
HUMAN\_PERSON                & HUMAN\_PERSON            & Thing           \\ \hline
GROUND\_ROAD                 & GROUND\_ROAD             & Stuff           \\
GROUND\_SPEED\_BUMP          & GROUND\_ROAD             & Stuff           \\
GROUND\_CURB                 & GROUND\_CURB             & Stuff           \\
GROUND\_PARKING              & GROUND\_PARKING          & Stuff           \\
GROUND\_BICYCLE\_LANE        & GROUND\_ROAD             & Stuff           \\
GROUND\_PEDESTRIAN\_CROSSING & GROUND\_ROAD             & Stuff           \\
GROUND\_ZEBRA\_ROAD\_MARKING & GROUND\_ROAD             & Stuff           \\
GROUND\_SIDEWALK             & GROUND\_SIDEWALK         & Stuff           \\
GROUND\_TERRAIN              & GROUND\_TERRAIN          & Stuff           \\
GROUND\_OTHER                & GROUND\_ROAD             & Stuff           \\ \hline
MANMADE\_BUILDING            & MANMADE\_BUILDING        & Stuff           \\
MANMADE\_FENCE               & MANMADE\_FENCE           & Stuff           \\
MANMADE\_OTHER               & MANMADE\_OTHER           & Stuff           \\
MANMADE\_POLE                & MANMADE\_POLE            & Thing           \\
MANMADE\_TRAFFIC\_SIGN       & MANMADE\_TRAFFIC\_SIGN   & Thing           \\
MANMADE\_TRAFFIC\_LIGHT      & MANMADE\_TRAFFIC\_LIGHT  & Thing           \\ \hline
NATURE\_VEGETATION           & NATURE\_VEGETATION       & Stuff           \\
NATURE\_TRUNK                & NATURE\_TRUNK            & Stuff           \\ \hline
DYNAMIC\_CAR                 & VEHICLE\_CAR             & Thing           \\
DYNAMIC\_BICYCLE             & VEHICLE\_BICYCLE         & Thing           \\
DYNAMIC\_PERSON              & HUMAN\_PERSON            & Thing           \\
DYNAMIC\_TRUCK               & VEHICLE\_TRUCK           & Thing           \\
DYNAMIC\_OTHER\_VEHICLE      & OTHER\_VEHICLE           & Thing           \\ \hline
\end{tabular}
\end{table*}
 
\begin{table*}[]
\centering
\captionsetup{justification=centering}
\caption{Navya3DSeg and SemanticKITTI mapping to common label kernel for cross-dataset generalization experiments}
\label{tab:cross_dataset_kernel}
\begin{tabular}{l|l|l}
\hline
\textbf{Navya3DSeg Label}                     & \textbf{SemanticKITTI Label} & \textbf{Cross-dataset generalization common labels} \\ \hline
UNKNOWN                                       & UNLABELED                    & UNKNOWN                                             \\ \cline{1-2}
OTHER                                         & OUTLIER                      & UNKNOWN                                             \\ \cline{1-2}
VEHICLE\_MOTORCYCLE                           & MOTORCYCLE                   & UNKNOWN                                             \\ \cline{1-2}
VEHICLE\_TRUCK                                & TRUCK                        & UNKNOWN                                             \\ \cline{1-2}
DYNAMIC\_VEHICLE\_TRUCK                       & MOVING\_TRUCK                & UNKNOWN                                             \\ \cline{1-2}
VEHICLE\_OTHER                                & OTHER\_VEHICLE               & UNKNOWN                                             \\ \cline{1-2}
DYNAMIC\_OTHER\_VEHICLE                       & MOVING\_OTHER\_VEHICLE       & UNKNOWN                                             \\ \cline{1-2}
GROUND\_OTHER                                 & OTHER\_GROUND                & UNKNOWN                                             \\ \cline{1-2}
MANMADE\_FENCE                                & FENCE                        & UNKNOWN                                             \\ \cline{1-2}
MANMADE\_OTHER                                & OTHER\_STRUCTURE             & UNKNOWN                                             \\ \cline{1-2}
MANDMADE\_OBJECT                              & OTHER\_OBJECT                & UNKNOWN                                             \\ \cline{1-2}
MANMADE\_TRAFFIC\_LIGHT                       & \cellcolor[HTML]{848482}     & UNKNOWN                                             \\ \cline{1-2}
\cellcolor[HTML]{848482}                      & BUS                          & UNKNOWN                                             \\ \cline{1-2}
\cellcolor[HTML]{848482}                      & ON-RAILS                     & UNKNOWN                                             \\ \cline{1-2}
\cellcolor[HTML]{848482}                      & MOVING\_BUS                  & UNKNOWN                                             \\ \cline{1-2}
\cellcolor[HTML]{848482}                      & MOVING\_ON-RAILS             & UNKNOWN                                             \\ \hline
VEHICLE\_CAR                                  & CAR                          & VEHICLE\_CAR                                        \\ \cline{1-2}
DYNAMIC\_CAR                                  & MOVING\_CAR                  & VEHICLE\_CAR                                        \\ \hline
VEHICLE\_BICYCLE                              & BICYCLE                      & VEHICLE\_BICYCLE                                    \\ \cline{1-2}
DYNAMIC\_BICYCLE                              & MOVING\_BICYCLE              & VEHICLE\_BICYCLE                                    \\ \hline
HUMAN\_PERSON                                 & PERSON                       & HUMAN\_PERSON                                       \\ \cline{1-2}
DYNAMIC\_PERSON                               & MOVING\_PERSON               & HUMAN\_PERSON                                       \\ \cline{1-2}
\cellcolor[HTML]{848482}                      & BICYCLIST                    & HUMAN\_PERSON                                       \\ \cline{1-2}
\cellcolor[HTML]{848482}                      & MOTORCYCLIST                 & HUMAN\_PERSON                                       \\ \cline{1-2}
\cellcolor[HTML]{848482}                      & MOVING\_BICYCLIST            & HUMAN\_PERSON                                       \\ \cline{1-2}
\cellcolor[HTML]{848482}                      & MOVING\_MOTORCYCLIST         & HUMAN\_PERSON                                       \\ \hline
GROUND\_ROAD                                  & ROAD                         & GROUND\_ROAD                                        \\ \cline{1-2}
GROUND\_SPEED\_BUMP                           & \cellcolor[HTML]{848482}     & GROUND\_ROAD                                        \\ \cline{1-2}
GROUND\_BICYCLE\_LANE                         & \cellcolor[HTML]{848482}     & GROUND\_ROAD                                        \\ \cline{1-2}
GROUND\_PEDESTRIAN\_CROSSING                  & \cellcolor[HTML]{848482}     & GROUND\_ROAD                                        \\ \cline{1-2}
GROUND\_ZEBRA\_ROAD\_MARKING                  & \cellcolor[HTML]{848482}     & GROUND\_ROAD                                        \\ \cline{1-2}
\cellcolor[HTML]{848482} & LANE\_MARKING                & GROUND\_ROAD                                        \\ \hline
GROUND\_SIDEWALK                              & SIDEWALK                     & GROUND\_SIDEWALK                                    \\ \cline{1-2}
GROUND\_CURB                                  & \cellcolor[HTML]{848482}     & GROUND\_SIDEWALK                                    \\ \hline
GROUND\_PARKING                               & PARKING                      & GROUND\_PARKING                                     \\ \hline
GROUND\_TERRAIN                               & TERRAIN                      & GROUND\_TERRAIN                                     \\ \hline
MANMADE\_BUILDING                             & BUILDING                     & MANMADE\_BUILDING                                   \\ \hline
NATURE\_VEGETATION                            & VEGETATION                   & NATURE\_VEGETATION                                  \\ \hline
NATURE\_TRUNK                                 & TRUNK                        & NATURE\_TRUNK                                       \\ \hline
MANMADE\_POLE                                 & POLE                         & MANMADE\_POLE                                       \\ \hline
MANMADE\_TRAFFIC\_SIGN                        & TRAFFIC\_SIGN                & MANMADE\_TRAFFIC\_SIGN                              \\ \hline
\end{tabular}
\end{table*}